\documentclass[10pt,twocolumn,letterpaper]{article}

\usepackage{cvpr}              
\usepackage{multirow}
\usepackage{colortbl} 
\usepackage{lineno}   
\definecolor{cvprblue}{rgb}{0.21,0.49,0.74}
\usepackage[pagebackref,breaklinks,colorlinks,allcolors=cvprblue]{hyperref}

\def\paperID{17598} 
\def\confName{CVPR}
\def\confYear{2025}

\title{CASP: Compression of Large Multimodal Models Based on Attention Sparsity}

\author{Mohsen Gholami, Mohammad Akbari, Kevin Cannons, and Yong Zhang \\
Huawei Technologies Canada Co., Ltd.\\
{\tt\small \{mohsen.gholami1, mohammad.akbari, kevin.cannons, yong.zhang3\}@huawei.com}}   
 
\begin{document}
\maketitle

\begin{abstract}
In this work, we propose an extreme compression technique for Large Multimodal Models (LMMs). While previous studies have explored quantization as an efficient post-training compression method for Large Language Models (LLMs), low-bit compression for multimodal models remains under-explored. The redundant nature of inputs in multimodal models results in a highly sparse attention matrix. We theoretically and experimentally demonstrate that the attention matrix's sparsity bounds the compression error of the Query and Key weight matrices. Based on this, we introduce CASP, a model compression technique for LMMs. Our approach performs a data-aware low-rank decomposition on the Query and Key weight matrix, followed by quantization across all layers based on an optimal bit allocation process. CASP is compatible with any quantization technique and enhances state-of-the-art 2-bit quantization methods (AQLM and QuIP\#) by an average of 21\% on image- and video-language benchmarks. The code is available \href{https://github.com/vbdi/casp}{here}\footnote{\url{https://github.com/vbdi/casp}}.
\end{abstract}

\section{Introduction}
Large multimodal Models (LMMs) have garnered significant attention in recent years due to their impressive performance in image- and video-language comprehension. Despite their substantial applications, LMMs are computationally expensive, which limits their broader use. For instance, 70B models like LLaVa-Onevision \cite{li2024llava} require 140GB of GPU memory to operate at 16-bit precision. Moreover, the
inference of LMMs requires significant electricity consumption, which raises concerns about environmentally friendly AI \cite{wang2024large}.
To address this problem, compression techniques such as knowledge distillation \cite{gholami-etal-2024-gold,MiniLLM}, quantization \cite{FrantarE23,lin2023awq,aqlm}, and low-rank factorization \cite{YuanZ23,WangX24} have been proposed.  

\begin{figure}[t!]
    \centering
    \includegraphics[width=0.9\linewidth]{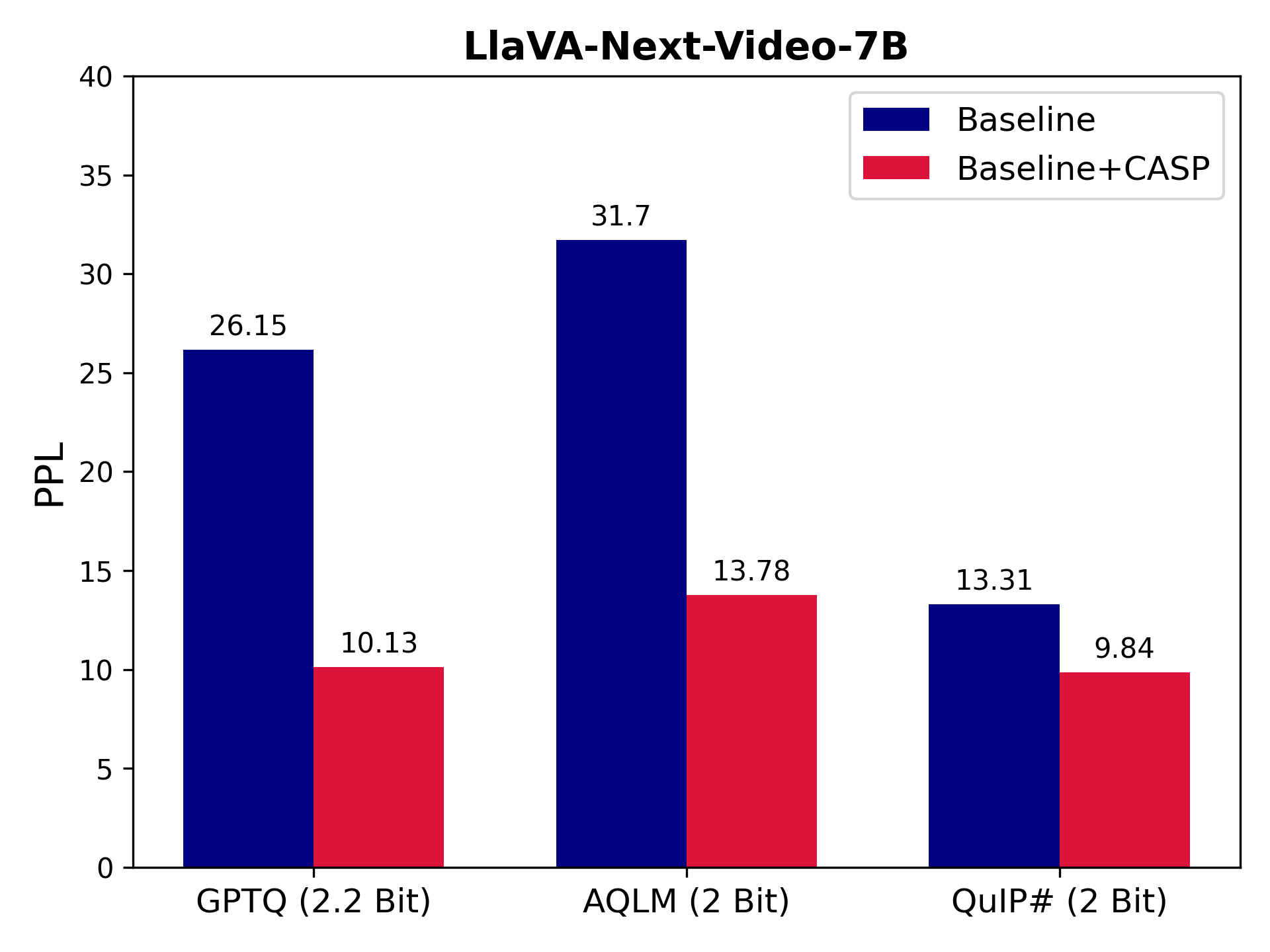}
    \vspace{-7pt}
    \caption{CASP considers the specific properties of LMMs and offers significant improvement over state-of-the-art model quantization methods. PPL: perplexity.}
    \vspace{-10pt}
    \label{fig:teaser}
\end{figure}

Recent studies predominantly focus on compressing large language models (LLMs), while the LMMs compression remains less explored. Although LLM compression techniques can be applied to LMMs, there are fundamental differences between the two. Popular LMMs convert visual modalities into hundreds to thousands of tokens, which are then concatenated with text tokens and fed to an underlying LLM. Unlike textual data, multimodal inputs are typically highly redundant. This redundancy is reflected in the attention layers, {which can effectively} be considered during {LMMs} compression.

Fig. \ref{fig:attention} illustrates the attention maps of LLaVa-Next-Video-7B compared to Llama2-7B for the same layer (i.e., layer 15). Despite LLaVa-Next-Video-7B utilizing Llama2-7B as its underlying LLM, the attention of the two models differs significantly. This observation raises two under-explored questions: 1) \textit{How do the activations and attentions of LMMs differ from those of LLMs?} 2) \textit{What insights should be considered to optimize LMMs compression?}

In LMMs, visual tokens typically receive much less attention compared to text tokens. 
For example, in LLaVa-1.5 \cite{liu2023llava}, vision tokens approximately receive ten times less attention compared with text tokens. 
This disparity is even more pronounced in LLaVa-Next-Video with greater number of visual tokens.
The cumulative attention on text tokens leads to a sparse attention map {(Fig. \ref{fig:attention}). We hypothesize that the sparse attention maps can be recovered by smaller attention weights, allowing for low-bit compression. This phenomenon is well-documented in the compressed sensing literature \cite{donoho2006most,candes2011probabilistic}, which shows that through optimization, a signal's sparsity can be leveraged to reconstruct it from far fewer samples than those required by the Nyquist–Shannon sampling theorem. While our problem differs from those addressed in compressed sensing, it is closely related. 

In addition, although existing post-training model compression methods nearly match the original model’s performance with up to 3-bit precision, lower bits (e.g., 2 bits) result in a significant accuracy drop. This effect remains an open problem for both LLMs and LMMs. AQLM \cite{aqlm} and QuIP\# \cite{quipsharp} are state-of-the-art techniques for 2-bit LLM quantization, but they both rely on time-consuming fine-tuning 
to restore the performance of compressed models. For instance, quantizing a 70B model on a single A100 GPU with AQLM can take 10-14 days. In this paper, we tackle this issue with a fine-tuning-free, post-training model compression approach for LMMs in the low-bit precision scenario. 
{To the best of our knowledge, this is the first study to explore 2-bit LMM compression.}

Specifically, we propose CASP, a model \textbf{C}ompression method for LMMs based on \textbf{A}ttention \textbf{SP}arsity. Motivated by our empirical observations, we theoretically  
show that the compression error of the attention weights ($W_q$ and $W_k$ in Eq. \ref{eq:att_mat}) is bounded by the sparsity of the attention maps. This implies that with sparser maps, greater compression of the attention weights can be achieved with negligible performance drop.
Consequently, we introduce a joint quantization and low-rank decomposition technique that compresses the attention weights to 6\% of their original size, equivalent to 1 bit. Initially, we perform low-rank decomposition on the attention weight matrices. Following this, we propose an optimal bit allocation strategy for each layer in the model to assign more bits to critical layers so that an average target bit rate can be achieved. Our method is orthogonal to any quantization technique. 
{CASP} significantly enhances the performance of state-of-the-art 2-bit quantization methods, including AQLM and QuIP\#, by an average of 35\% and 7\% on different image- and video-language benchmarks. 
Additionally, we demonstrate that CASP can be applied to LLMs, improving AQLM and QuIP\# by an average of 11\% and 2.7\% on language-only benchmarks.

\textbf{Contributions.} Our major contributions are as follows:
\begin{enumerate}
     \item Providing both theoretical and experimental insights showing the compression error of the attention weight matrices is bounded by the {attention maps} sparsity.
     \item Proposing CASP, a novel low-bit LMM compression method based on attention sparsity.
    \item {CASP is a finetuning-free approach, that is compatible with any quantization method and also applicable to both LMMs and LLMs.}
    \item Validating CASP’s performance across a wide range of image-language, video-language, and language-only benchmarks using 5 different LMMs and LLMs, achieving state-of-the-art results.
\end{enumerate}

\section{Related Works}
LLM compression methods fall into two main categories: training-aware and post-training. Training-aware methods, like quantization aware training (QAT) \cite{NagelM22,XiH23}, integrate quantization with network training, making models more suitable for quantization. However, due to the high time and computational costs of training modern LLMs, these methods are less favored. Post-training compression methods are more practical as they compress pre-trained models in one-shot without additional training \cite{EgiazarianV24,FrantarE23,MaX23,quipsharp,WangX24,LinC24}.

\begin{figure*}[t!]
    \centering
    \includegraphics[width=0.9\linewidth]{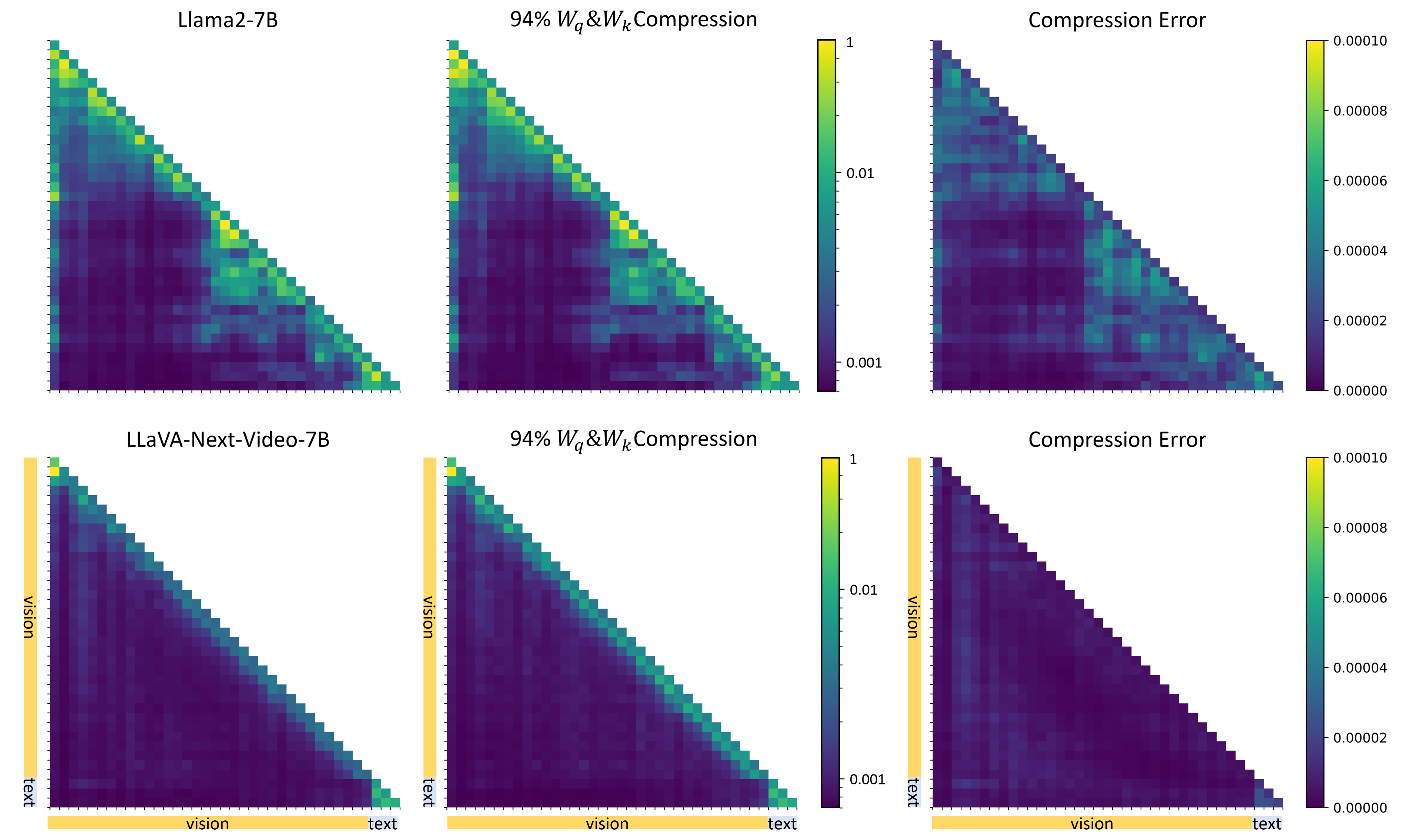}
    \vspace{-5pt}
    \caption{Left column: Comparison of LLaVa-Next-Video-7B and Llama2-7B attention maps ($S$ in Eq. \ref{eq:att_mat}) at Layer 15 . Despite LLaVa-Next-Video using Llama2 as its base LLM, there is a notable difference in their maps, with LLaVa showing high sparsity. Middle column: Attention maps when $W_q$ and $W_k$ (i.e., attention weights) are 94\% compressed (equivalent to 1 bit). Right column: Compression errors ($E$ in Eq. \ref{eq:error}). The sparsity in LLaVa's map results in smaller errors when compressing $W_q$ and $W_k$. }
    \vspace{-8pt}
    \label{fig:attention}
\end{figure*}

In post-training LLM compression, the main methods are quantization \cite{FrantarE23,EgiazarianV24,CheeJ23,quipsharp}, pruning \cite{FrantarE23,MaX23}, knowledge distillation \cite{HsiehC23,GuY24}, and low-rank decomposition \cite{HsuY22,YuanZ23,WangX24,LinC24}. Post-training quantization (PTQ) is particularly notable for its efficiency, as it quantizes network parameters after training, requiring less computation than quantization aware training (QAT) while still achieving competitive performance \cite{EgiazarianV24,quipsharp}.

Early works on PTQ existed before the rise of LLMs \cite{NagelM20,GholamiA21}. Initial PTQ methods for LLMs used round-to-nearest projections, adjustable for different memory/accuracy needs \cite{YaoZ22,DettmersT22,ParkG22}. GPTQ \cite{FrantarE23} introduced a data-aware approach, minimizing $l_2$ errors on {a calibration dataset} for each network layer using a large-scale solver. 
Recent methods like QuIP \cite{CheeJ23} and QuIP\# \cite{quipsharp} use a two-step PTQ process involving weight smoothing and subsequently mapping weights onto a lattice codebook. AQLM \cite{EgiazarianV24}, similar to QuIP, uses a data-driven codebook but with additive weight encoding and omitting the smoothing step. PTQ has not yet been applied to LMMs, which will be addressed by our proposed solution.

Another post-training model compression method is low-rank decomposition, such as singular value decomposition (SVD), which approximates matrices with lower-rank matrices to compress the model \cite{HsuY22,YuanZ23,WangX24,LinC24}. Despite its potential, SVD for LLM compression is relatively unexplored. ASVD \cite{YuanZ23} was one of the first approaches but suffered from performance degradation at high compression ratios. SVD-LLM \cite{WangX24} improved model accuracy at high compression by using a data whitening strategy to account for the impact of each singular value on the compression loss. MoDeGPT \cite{LinC24} uses a different strategy by grouping matrices into larger modules and applying 3 types of module-dependent matrix approximations. 
Despite these initial efforts, it appears that such methods have not been considered for LMMs.

The literature has mainly focused on applying single post-training compression methods to pre-trained LLMs. However, pruning, quantization, and low-rank decomposition can be combined. LQ-LoRA \cite{GuoP24} is a recent work that combines these strategies by performing matrix decomposition to create a high precision low-rank component and a memory-efficient quantized component. No prior work has applied PTQ, low-rank decomposition, or their combination to LMMs, which is the focus of our proposed approach.

\section{Method}

In this section, we first describe the attention mechanism, its sparsity, and low-rank features, which motivate us to perform low-rank decomposition on the attention weights. Following that, we prove the theorem that the compression error is bounded by the attention map sparsity. Next, we explain the low-bit quantization process as the second phase of our method, followed by a proof showing that our bit allocation is optimal.

\subsection{Attention Weights Compression}
\label{ssec:attention}

The scaled-dot-product (SDP) attention in transformer-based models \cite{vaswani2023attentionneed} operates on queries \( \mathbf{Q} = X W_q \in \mathbb{R}^{N \times d} \) and keys \( \mathbf{K} = X W_k \in \mathbb{R}^{N \times d} \), where \( X \) represents the input activations from the previous layer, \( W_q \) and \( W_k \) are the model weights, \( d \) is the hidden dimension, and \( N \) is the number of tokens. The attention maps are then defined as:
\vspace{-5pt}
\begin{equation}
     \mathbf{S} = \text{Softmax}\left(\frac{X W_q W_k^\top X^\top}{\sqrt{d}}\right) =\text{Softmax}(Y),
     \label{eq:att_mat}
\end{equation}
where \( \mathbf{S} \in \mathbb{R}^{N \times N} \). This operation has three specific criteria that are crucial in the context of LMM compression. \textit{First}, unlike other weight matrices in transformers (such as those in MLPs, attention-value, and attention-output layers), \( W_q \) and \( W_k \) exhibit a highly low-rank structure \cite{YuanZ23}. 
\textit{Second}, while activation outputs in transformers are generally not sparse, the attention map \( \mathbf{S} \), representing the activation output of the layer, is sparse. {\textit{Third}, the sparsity of \( \mathbf{S} \) is even more pronounced in LMMs because vision tokens receive significantly less attention.}

{Given the above motivations, we perform a data-aware low-rank decomposition on $W_q$ and $W_k$ to obtain two low-rank weight matrices $W'_q=A_qB_q$ and $W'_k=A_kB_k$.  
First, following \cite{WangX24}, we compute the covariance matrix \( \mathbf{C} \) of the calibration data and perform Cholesky decomposition to obtain the lower triangular matrix \( \mathbf{L} \). The whitening matrix \( \mathcal{A} \) is then derived as \( \mathbf{L}^{-1} \). By applying \( \mathcal{A} \) to the calibration data, we transform the data into a whitened space. Subsequently, we perform low-rank decomposition on the whitened data to obtain \( W’ \).} The approximated attention map is then calculated as:
\vspace{-5pt}
\begin{equation}
     \mathbf{S'} = \text{Softmax}\left(\frac{X {W'}_q {W'}_k^\top X^\top}{\sqrt{d}}\right) =\text{Softmax}(Y').
     \vspace{-5pt}
     \label{eq:98}
\end{equation}

The error of the low-rank approximation is calculated using the Forbenius norm of the difference between the original and approximated attention maps as follows: 
\vspace{-3pt}
\begin{equation}
     {E} = || \mathbf{S}' - \mathbf{S}||=||\text{Softmax}(Y')-\text{Softmax}(Y)||.
     \vspace{-3pt}
    \label{eq:error}
\end{equation}

Let $\mathcal{S}$ denote the attention map sparsity, calculated as the proportion of elements in $\mathbf{S}$ {with values above a small threshold $\eta$,} divided by the total number of elements in $\mathbf{S}$. In the following theorem, we demonstrate that the attention compression error is bounded by the sparsity of the attention map. Consequently, the compression error decreases as the attention map becomes sparser.

\subsubsection{Theorem and Proof}

{\textbf{Theorem 1.} \textit{The compression error of the $W_q$ and $W_k$ (Eq. \ref{eq:error}) is bounded by the sparsity of the attention map $\mathbf{S}$ (Eq. \ref{eq:att_mat}).}}

\noindent\textbf{Proof.} 
The compression error can be written as: 
\vspace{-5pt}
\begin{equation}
    E =\| \text{Softmax}(Y + \delta Y) - \text{Softmax}(Y) \|,
    \label{eq:100}
\end{equation} 
where 
\( Y, \delta Y \in \mathbb{R}^{N \times N} \) and \( \| \cdot \| \) denotes the Frobenius norm. Since Softmax is applied to each row of $Y$ independently, we can write the above equation as: 
\vspace{-5pt}
\begin{equation}
    E =\sum_{i=1}^{N} \| \text{Softmax}(Y_i + \delta Y_i) - \text{Softmax}(Y_i) \|,
    \label{eq:101}
\end{equation}
where $Y_i \in R^{N \times 1}$ and $\delta Y_i \in R^{N \times 1}$ are the $i$th row of $Y_i$ and $\delta Y_i$, respectively.
The difference between the Softmax of $(Y_i+\delta Y_i)$ and $(Y_i)$ can be approximated using a first-order Taylor expansion around $(Y)$:
\vspace{-5pt}
\begin{equation}
    \text{Softmax}(Y_i + \delta Y_i) \approx \text{Softmax}(Y_i) + \nabla \text{Softmax}(Y_i) \cdot \delta Y_i,
    \label{eq:102}
\end{equation} 
where \( \nabla \text{Softmax}(Y_i) \in R^{ N \times N}\) is the Jacobian matrix of the Softmax function. Substituting this approximation into Eq. \ref{eq:101}, we get: 
\vspace{-5pt}
\begin{equation}
    E =\sum \| \nabla \text{Softmax}(Y_i) \cdot \delta Y_i \| + \epsilon_i.
    \label{eq:103}
\end{equation} 
where \( \epsilon_i \) represents the approximation error introduced by the first-order Taylor expansion. This simplification shows that \( E \) is approximately equal to the Frobenius norm of the product between the Jacobian of the Softmax function and the perturbation \( \delta Y_i \). Given the above equation, we can define a lower bound for \( E \):
\vspace{-5pt}
\begin{equation}
    E \leq \sum \| \nabla \text{Softmax}(Y_i)\| \cdot \| \delta Y_i \| + \epsilon_i.
    \label{eq:104}
\end{equation} 

\begin{figure}[t!]
    \centering
    \includegraphics[width=1\linewidth]{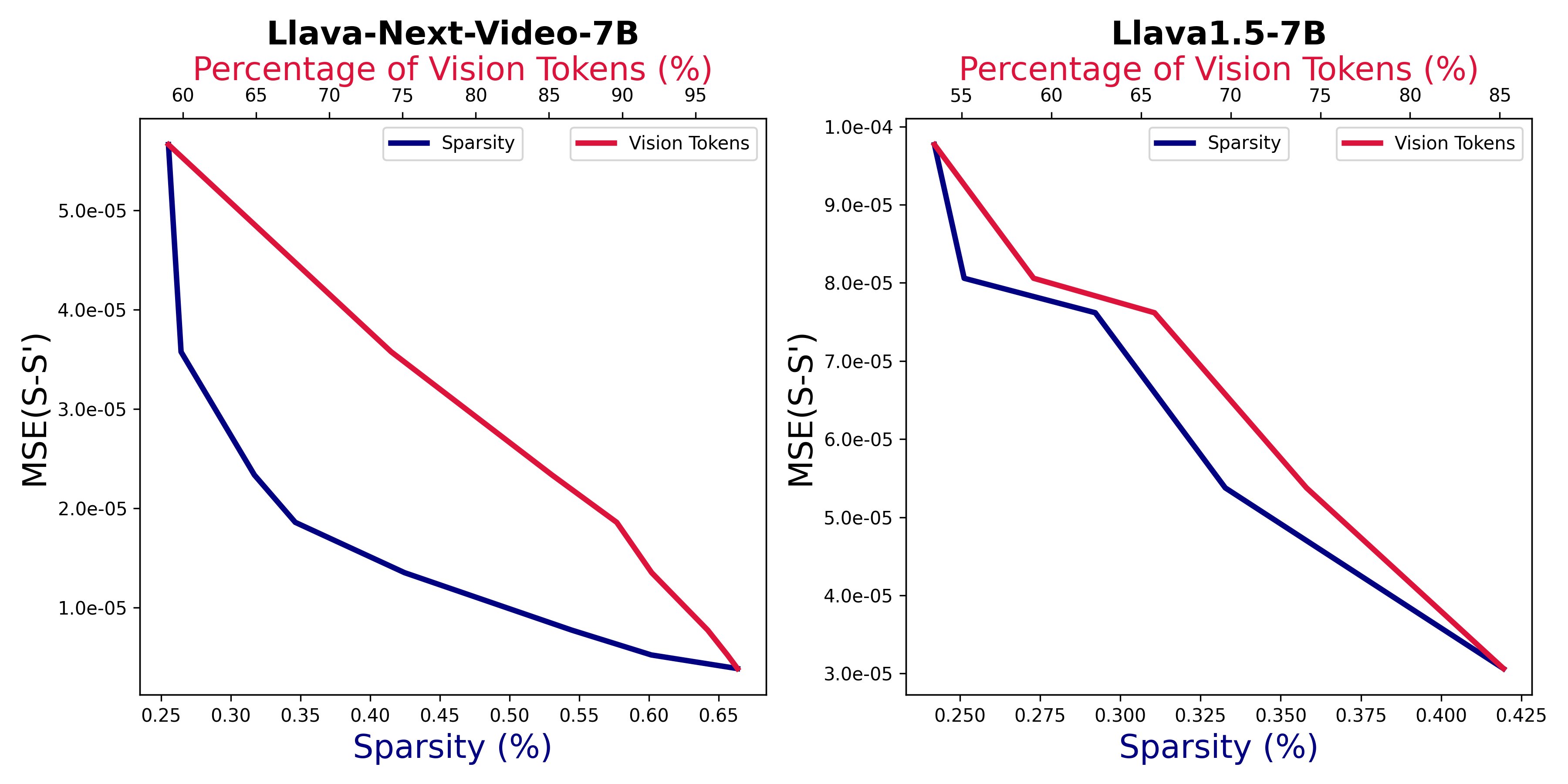}
    \vspace{-15pt}
    \caption{Compression error $E$ (Eq. \ref{eq:error}) for LLaVa-Next-Video and  LLaVa1.5 decreases when the percentage of visual tokens increases (i.e., more sparse attention map). 
    }
    \vspace{-5pt}
    \label{fig:proof2}
\end{figure}

The Jacobian matrix \( J=\nabla \text{Softmax}(Y_i) \) of the Softmax function is an \( N \times N \) matrix where each element \( J_{jm} \) is the partial derivative of the \( j \)th output with respect to the \( m \)th input. The elements of the Jacobian are: 
\vspace{-3pt}
\begin{equation}
   J_{jm} = \frac{\partial \text{Softmax}({Y_i})_j}{\partial z_m}, 
   \vspace{-3pt}
   \label{eq:105}
\end{equation}
which can be expressed as:
\vspace{-3pt}
\begin{equation}
    J_{jm} = \text{Softmax}({Y_i})_j \big(\delta_{jk} - \text{Softmax}({Y_i})_j\big),
    \vspace{-3pt}
    \label{eq:106}
\end{equation}
where \( \delta_{jm} \) is the Kronecker delta, which is 1 if \( j = m \) and 0 otherwise. In matrix form, the Jacobian \( J \) can be written as: 
$J = \text{diag}(\mathbf{z}) - \mathbf{z} \mathbf{z}^\top$,
where \( \mathbf{z} = \text{Softmax}({Y_i}) \) and \( \text{diag}(\mathbf{z}) \) is a diagonal matrix with the elements of \( \mathbf{z} \) on the diagonal. 
Define the density \( \mathcal{D} \) of \( Y \) as the proportion of non-zero elements to the total number of elements, in which case the norm of the Jacobian \( J \) can be approximated as: $||J|| \approx (1-\frac{1}{N\mathcal{D}})^2$. Therefore, based on Eq. \ref{eq:104}, we have:
\vspace{-3pt}
\begin{equation}
\begin{split}
   E \leq & (1-\frac{1}{N\mathcal{D}})^2  \sum \| \delta Y_i \| + \epsilon_i \\
  \implies E \leq& (1-\frac{1}{N\mathcal{D}})^2  \| \delta Y \| + \epsilon.
  \label{eq:upperbound}
\end{split}
\end{equation}
Therefore, the upper bound of compression error $E$ decreases as the density  $\mathcal{D}$  decreases, which corresponds to an increase in sparsity  $\mathcal{S}$. 
{Note that the proof is intended to be general and applies to any approximated attention map.}
Moreover, in the context of low-rank compression, as the sparsity of attention map increases (i.e., smaller $\mathcal{D}$), we can have lower ranks and truncate more eigenvalues from $W_q$ and $W_k$, while maintaining the same upper bound for $E$ in Eq. \ref{eq:upperbound}

Fig. \ref{fig:proof2} shows that sparser attention maps (i.e., higher ratio of vision tokens) result in lower error \( E \) in both LLaVA-Next-Video and LLaVA-1.5. This suggest that the empirical evidence is aligned with the theoretical insight in Eq. \ref{eq:upperbound}. 

\textbf{}
\subsection{Quantization with Optimal Bit Allocation}
\label{ssec:optimal}
PTQ methods minimize the layer-wise reconstruction $||XW-X\hat{W}||$, where $\hat{W}$ is the quantized weight matrix and $X$ is the activation. 
State-of-the-art low-bit PTQ methods \cite{aqlm,quipsharp} use vector quantization and quantize a group of $g$ weights as a  $g$ dimensional vector. 
In $n$-bit vector quantization, a vector is quantized to one of  $2^{ng}$  vectors in  $\mathbb{R}^g$ , forming a  $2^{ng} \times g$  codebook $C$ \cite{aqlm,quipsharp}.

As we compress the model by low-rank factorization of $W_q$ and $W_k$ matrices, we can quantize important layers to higher bits and obtain the same average quantization bits as uniform quantization. We adopt the Block Influence score \cite{men2024shortgpt} to measure the importance of each layer as:
\vspace{-5pt}
\begin{equation}
    {s_l}=1-\mathbb{E}\frac{X_\text{in}^\top X_{\text{out}}}{||X_\text{in}||||X_\text{out}||},
\end{equation}
where $X_\text{in}$/$X_\text{in}$ are input/output activations of layer $l$. We then try to maximize the sum of importance scores, weighted by the number of parameters remaining in each layer. 
{To mitigate the adverse effects of excessive sparsity, we utilize entropic regularization to achieve smoothing.}
The formulation of this constrained optimization problem is as follows: 
\vspace{-5pt}
\begin{equation}
    \max_{b_1:b_L} \sum_{l=1}^{L} s_l b_l p_l + \mu\sum_{l=1}^{L} H(b_l),     
    \text{~~~s.t.~~~}
    \frac{1}{P}\sum_{l=1}^{L} b_l p_l  = B_\text{avg},
    \label{eq:quant1}
\end{equation}
where $ H(b_l) = -b_l \log(b_l) $ is the entropy, $\mu$ is the regularization parameter, and $L$ is the number of layers of the model. $b_l$ and $p_l$ are the quantization bits and the number of parameters after low-rank decomposition of $l\text{th}$ layer of the model. Also, $P$ is the total number of parameter of the original model and $B_\text{avg}$ is the target required quantization bit. 
The optimal layer bit distribution can be computed as:
\vspace{-5pt}
\begin{equation}
    b_l = \frac{1}{p_l}P B_\text{avg} \times \text{Softmax} (s_l p_l/\mu),
    \label{eq:quant2}
\end{equation}
\subsubsection{Theorem and Proof}
\textbf{Theorem 2.} \textit{Given the objective function in Eq. \ref{eq:quant1}, the bit allocation in Eq. \ref{eq:quant2} is optimal.}

\noindent\textbf{Proof.} The Lagrangian for the constrained optimization problem in Eq. \ref{eq:quant1} can be written as:
\vspace{-5pt}
\begin{equation}
\mathcal{L}(b, \lambda) = \sum_{l=1}^{L} \big( s_l b_l p_l + \mu H(b_l) \big) + \lambda \left( B_\text{avg} - \sum_{l=1}^{L} \frac{b_l p_l}{P}\right),
\label{eq:quant3}
\end{equation}
where \( \lambda \) is the Lagrange multiplier.
To find the optimal \( b_l \), we take the partial derivative of the Lagrangian with respect to \( b_l \) and set it to zero:
\vspace{-5pt}
\begin{equation}
\frac{\partial \mathcal{L}}{\partial b_l} = s_l p_l - \mu\log(b_l) - \mu - \lambda \frac{p_l}{P} = 0.
\label{eq:quant3}
\end{equation}
Solving for \( b_l \):
\vspace{-5pt}
\begin{equation}
    b_l = \exp \left(  \frac{s_l p_l - \mu - \lambda \frac{p_l}{P}} {\mu} \right).
\label{eq:quant4}
\end{equation}
After substituting $b_l$ into the constraint in Eq. \ref{eq:quant1}:
\vspace{-5pt}
\begin{equation}
   B_\text{avg} = \sum_{l=1}^{L} \exp \left(  \frac{s_l p_l - \mu - \lambda \frac{p_l}{P}} {\mu} \right) \frac{p_l}{P}.  
\label{eq:quant5}
\end{equation}
Therefore, having Eq. \ref{eq:quant4} and \ref{eq:quant5}, we can calculate $b_l$ by:
\vspace{-5pt}
\begin{equation}
    b_l = \frac{B_\text{avg} P}{p_l} \text{ Softmax}\left(\frac{s_l p_l - \mu - \lambda \frac{p_l}{P}}{\mu}\right).
\end{equation}

Since we used the same rank for the low-rank decomposition of $W_q$ and $W_k$ in different layers, $p_l$ is constant. As the Softmax function is invariant to adding constant values,  we can simplify the above equation to $b_l = \frac{P B_\text{avg}}{p_l} \times \text{Softmax} (s_l p_l/\mu)$ that is given in Eq. \ref{eq:quant2}.

\setlength{\tabcolsep}{4pt} 
\renewcommand{\arraystretch}{0.84}
\begin{table*}[]
    \centering
    \footnotesize
    \begin{tabular}{c|l|c|ccc|ccccc|c}
          \toprule
         && &{LiveB}&{LWilder}&{LCOCO}&{SEEDB}&{SQA}&{MMMU}&{MMB}&{MME} & \textcolor{blue}{{Avg. Rel.}}   \\
         &{Method}& Bit& (PPL$\downarrow$) & (PPL$\downarrow$) & (PPL$\downarrow$) &  {(Acc$\uparrow$)} & {(EM$\uparrow$)} & {(Acc$\uparrow$)} & {(Acc$\uparrow$)} & (Cognition$\uparrow$) & \textcolor{blue}{{Improv.}}\\ \midrule 
        \multirow{7}{*}{\rotatebox[origin=c]{90}{\textcolor{NavyBlue}{\textbf{LLaVA1.5-7B}}}}&  \cellcolor[gray]{0.9}Original  & \cellcolor[gray]{0.9}16 & \cellcolor[gray]{0.9}5.5&\cellcolor[gray]{0.9}4.2&\cellcolor[gray]{0.9}4.5&\cellcolor[gray]{0.9}65.4&\cellcolor[gray]{0.9}67.9 & \cellcolor[gray]{0.9}35.6&\cellcolor[gray]{0.9}62.9&\cellcolor[gray]{0.9}323& \\ \cmidrule(lr){2-12}
        &GPTQ & 2.2 & 38.1&138.7&7.5&3.0&9.17&23.7&4.6&175\\
        &CASP\textsubscript{GPTQ} & 2.2&\textbf{8.2}&\textbf{10.5}&\textbf{5.6}&\textbf{24.2}&\textbf{43.9}&\textbf{25.7}&\textbf{22.5}&\textbf{231} & \textcolor{blue}{+125.6\%}\\ \cmidrule(lr){2-12}
        &AQLM & 2 & 14.9& 23.3&10.8&41.7&36.7&24.5&25.6&241\\
        &CASP\textsubscript{AQLM} & 2& \textbf{7.9}&\textbf{8.2}&\textbf{5.7} &\textbf{52.2}&\textbf{50.8}&\textbf{27.4} &\textbf{38.9}&\textbf{298} & \textcolor{blue}{+38.7\%}\\ \cmidrule(lr){2-12}
        &QuIP\# & 2 & 9.2&6.7&5.3&61.4&60.3&29.5&52.3&243\\ 
        &CASP\textsubscript{QuIP\#}& 2 & \textbf{7.1} &\textbf{6.6} & \textbf{5.3} &\textbf{63.1} &\textbf{61.6}&\textbf{32.1}&51.8&\textbf{263}  & \textcolor{blue}{+5.7\%} \\ \cmidrule(lr){1-12}
        \multirow{7}{*}{\rotatebox[origin=c]{90}{\textcolor{NavyBlue}{\textbf{LLaVA1.5-13B}}}} &\cellcolor[gray]{0.9}Original  & \cellcolor[gray]{0.9}16 &\cellcolor[gray]{0.9}5.1 &\cellcolor[gray]{0.9}4.0 & \cellcolor[gray]{0.9}4.2 & \cellcolor[gray]{0.9}67.7 & \cellcolor[gray]{0.9}71.6 & \cellcolor[gray]{0.9}36.5 & \cellcolor[gray]{0.9}68.1 & \cellcolor[gray]{0.9}312\\ \cmidrule(lr){2-12}
        &GPTQ & 2.2 &13.7&36.2&\textbf{5.5} & 52.9 & 24.6 & 26.5 & 32.5 & 247\\
        &CASP\textsubscript{GPTQ} & 2.2&\textbf{8.6}&\textbf{16.0}&6.0 & \textbf{59.7} & \textbf{59.7} & \textbf{29.7} & \textbf{51.5} & \textbf{304} &\textcolor{blue}{+40.0\%} \\ \cmidrule(lr){2-12}
        &AQLM & 2 & 10.2&18.1& 7.9 &56.4&57.7&28.2& 38.3 & 207 \\
        &CASP\textsubscript{AQLM} & 2& \textbf{6.2} & \textbf{7.3} & \textbf{5.6}&\textbf{64.4}&\textbf{67.9}&\textbf{33.1}&\textbf{58.7}&\textbf{261} & \textcolor{blue}{+32.0\%}\\ \cmidrule(lr){2-12}
        &QuIP\# & 2 &6.0&6.1&4.7&66.7&68.3&33.8&\textbf{63.4}&270\\ 
        &CASP\textsubscript{QuIP\#}& 2 & \textbf{6.0}&\textbf{5.4}&\textbf{4.7} &\textbf{66.7}&\textbf{71.2}&\textbf{33.8}&62.6&\textbf{293} & \textcolor{blue}{+2.7\%} \\ \cmidrule(lr){1-12}
        \multirow{7}{*}{\rotatebox[origin=c]{90}{\textcolor{NavyBlue}{\textbf{LLaVA-Next-7B}}}}&\cellcolor[gray]{0.9}Original  & \cellcolor[gray]{0.9}16 & \cellcolor[gray]{0.9}6.0 & \cellcolor[gray]{0.9}3.8& \cellcolor[gray]{0.9}4.9&\cellcolor[gray]{0.9}69.9&\cellcolor[gray]{0.9}70.1&\cellcolor[gray]{0.9}36.1&\cellcolor[gray]{0.9}66.9&\cellcolor[gray]{0.9}312   \\ \cmidrule(lr){2-12}
        &GPTQ & 2.2 & 29.9 & 211.0&88.7 & 5.9 & 9.42 & 25.1 & 9.36 & 151\\
        &CASP\textsubscript{GPTQ} & 2.2& \textbf{10.6}&\textbf{10.5}&\textbf{6.3} & \textbf{37.4} & \textbf{37.5} & \textbf{27.1} & \textbf{10.0} &\textbf{199}& \textcolor{blue}{+141.3\%} \\ \cmidrule(lr){2-12}
        &AQLM & 2 & 19.3 & 26.6 & 9.5&30.3&32.7&24.4&12.2 &162 \\
        &CASP\textsubscript{AQLM} & 2& \textbf{9.3} & \textbf{7.7} & \textbf{6.4} & \textbf{60.7} &\textbf{53.5} & \textbf{29.5}  & \textbf{43.3} & \textbf{219} & \textcolor{blue}{+78.6\%}\\ \cmidrule(lr){2-12}
        &QuIP\# & 2 & 7.7 & 5.6& 5.3& 65.6&60.6&31.2&\textbf{56.7}&\textbf{263} \\ 
        &CASP\textsubscript{QuIP\#}& 2 & \textbf{7.3} & \textbf{4.8} & \textbf{5.1}&\textbf{66.4}&\textbf{61.1}&\textbf{32.4}&55.4&241 & \textcolor{blue}{+2.3\%} \\
    \bottomrule
    \end{tabular}
    \vspace{-3pt}
    \caption{
    {Comparison results of our proposed CASP and different baselines (including GPTQ, AQLM, and QuIP\#) on LLaVA models across image-language understanding datasets. All the results are in 2-bit precision, except GPTQ (i.e., 2.2-bit). The average relative improvement of CASP over the baselines is also provided. 
    }
    }    
    \vspace{-10pt}
    \label{tab:image-bench}
\end{table*}

\section{Experiments}
In this section, we analyze the performance of the proposed method compared to baselines across different image-language, video-language, and language-only models and benchmarks. The ablation studies consider the effects of the calibration dataset type, higher bit precision quantization, optimal bit allocation, and also the impact of the ratio of text vs. vision tokens.

\subsection{Experimental Setting}

\textbf{Models.} For the experiments on image-language benchmarks, we use three different LMMs namely LLaVA1.5-7B\footnote{https://huggingface.co/llava-hf/llava-1.5-7b-hf} \cite{liu2023llava}, LLaVA1.5-13B\footnote{https://huggingface.co/llava-hf/llava-1.5-13b-hf}, and LLaVA-Next-7B\footnote{https://huggingface.co/llava-hf/llava-v1.6-vicuna-7b-hf} (i.e., LLaVA1.6-7B) \cite{liu2023improved}. We also use LLaVA-Next-Video-7B\footnote{https://huggingface.co/llava-hf/LLaVA-NeXT-Video-7B-hf}
 \cite{zhang2024llavanextvideo} for the video-language experiments. All the above-mentioned models use Llama2-7B \cite{touvron2023llama} or 13B as their underlying LLM (depending on the model size) and CLIP \cite{radford2021learning} as their vision encoder. LLaVA1.5 and LLaVA-Next encode the input image to 576 and a dynamic number of visual tokens, respectively, while LLaVA-Next-Video uses 144 visual tokens for each frame in the video.

\noindent\textbf{Metrics and Benchmarks.} 
Our main evaluation metric is perplexity (PPL), which is the metric commonly used for quantization methods in the literature \cite{FrantarE23,quipsharp,lin2023awq}. PPL is defined as the exponentiation of the average negative log-likelihood of a sequence of tokens \cite{Jelinek1977PerplexityaMO}. We further evaluate CASP on different downstream tasks that use their specific metrics. Except for the PPL results, all other results on downstream datasets are obtained using the LMMs-eval framework\footnote{https://github.com/EvolvingLMMs-Lab/lmms-eval/tree/v0.2.0} \cite{zhang2024lmmsevalrealitycheckevaluation}. 
In order to measure PPL, we use LiveBench {(LiveB)} \cite{livebench} and LLaVA-Bench-Wilder {(LWilder)} \cite{Jelinek1977PerplexityaMO} as open-ended QA datasets and LLaVA-Bench-COCO (LCOCO) \cite{Jelinek1977PerplexityaMO} as an image-captioning dataset. For downstream task performance analysis, multi-choice QA benchmarks such as SEED-Bench \cite{seedbench}, MMU \cite{mmu}, ScienceQA  (SQA) \cite{scieniceqa}, MME \cite{mme}, and MMBench \cite{mmbnech} benchmarks are used. 
For video-language tasks, VideoDetailCaption \cite{Jelinek1977PerplexityaMO} and VideoChatGPT (temporal) \cite{videochatgptT} are used as representative open-ended video QA datasets, which are evaluated using PPL, ROUGE-L (RG-L), and OpenAI's GPT score with GPT-4o-mini (i.e., out of 5) \cite{gpt4omini}. 
Moreover, we perform experiments over three multiple-choice QA benchmarks, NextQA \cite{nextqa}, VideoMME (VMME) \cite{fu2024video}, and ActivityNetQA \cite{activitynetqa}, which are evaluated in terms of Exact Match (EM) or accuracy. 

\noindent\textbf{Baselines.} We use 3 recent quantization methods, GPTQ \cite{FrantarE23}, AQLM \cite{aqlm}, and QuIP\# \cite{quipsharp}, as our baselines. 
GPTQ is designed for the quantization of 3-bit or higher. For 2-bit, GPTQ employs a group size of 128, which is equivalent to 2.2-bit. 
AQLM offers three different schemes of ``Number of Codebooks'' $\times$ ``Codebook Size'' for low-bit compression: $1 \times 16$, $8 \times 8$, and $1 \times 8$. 
Among these, only $1 \times 8$ is equivalent to 2-bit quantization, which we utilize in this paper. 
QuIP\# uses the E8P12 codebook for 2-bit quantization. Although both QuIP\# and AQLM use time-consuming fine-tuning, for a fair comparison with our method, we report the results without fine-tuning. In all the experiments in this paper including the baselines and our method (for both low-rank factorization and quantization), we use a calibration dataset consisting of 1024 samples from RedPajama \cite{together2023redpajama}, each with a sequence length of 4096. 

\noindent\textbf{Configuration.} $W_q$ and $W_k$ are low-rank compressed to 25\% of their original size for GPTQ and AQLM (i.e., 75\% of eigenvalues are removed). For QuIP\#, this number is 50\%. 3-bit quantization is then performed for more important layers chosen by the optimal bit allocation procedure in Section \ref{ssec:optimal}. For the rest of the layers, 2-bit is used.

\renewcommand{\arraystretch}{1}
\setlength{\tabcolsep}{3pt} 
\begin{table*}[t!]
    \centering
    \footnotesize
    \begin{tabular}{l|l|c|ccc|ccc|ccc|c}
          \toprule
         && &\multicolumn{3}{c|}{VideoChatGPT}&\multicolumn{3}{c|}{VideoDetailCaption }& ActivityNet & NextQA & VMME & \textcolor{blue}{{Avg. Rel.}}  \\
         & {Method}& Bit& (RG-L$\uparrow$) & (PPL$\downarrow$) & (Score$\uparrow$) & (RG-L$\uparrow$) & (PPL$\downarrow$) & (Score$\uparrow$)  & (Score$\uparrow$ $|$ Acc$\uparrow$) & (EM$\uparrow$) & (Acc$\uparrow$) & \textcolor{blue}{{Improv.}}  
         \\ \cmidrule(lr){1-13}
        \multirow{7}{*}{\rotatebox[origin=c]{90}{\textcolor{NavyBlue}{\textbf{LLaVA-Next-Video-7B}}}}&\cellcolor[gray]{0.9}Original  & \cellcolor[gray]{0.9}16 & \cellcolor[gray]{0.9}0.280&\cellcolor[gray]{0.9}7.1 & \cellcolor[gray]{0.9}1.76 &\cellcolor[gray]{0.9}0.239&\cellcolor[gray]{0.9}6.9 & \cellcolor[gray]{0.9}2.60 &\cellcolor[gray]{0.9}2.58 $|$ 46.36
        &\cellcolor[gray]{0.9}56.61 & \cellcolor[gray]{0.9}35.63 & \cellcolor[gray]{0.9} \\ \cmidrule(lr){2-13}
        &GPTQ & 2.2 & 0.196 &26.2&0.40 & 0.206&19.1&0.52&1.00 $|$ 15.02 & 21.50 & 12.74 & \\
        &CASP\textsubscript{GPTQ} & 2.2& \textbf{0.258} & \textbf{10.0}&\textbf{0.68} &\textbf{0.251}&\textbf{9.5}&\textbf{1.15}&\textbf{2.20 $|$ 38.30} &   \textbf{25.48} & \textbf{19.41} & \textcolor{blue}{+69.8\%}\\ 
        \cmidrule(lr){2-13}
        &AQLM & 2 &  0.198 & 31.7&{0.48} &\textbf{0.245}&24.3&{0.42}&{1.13} $|$ {22.54} & 19.83 & {11.70} &  \\
        &CASP\textsubscript{AQLM} & 2&\textbf{0.241}&\textbf{13.8}& \textbf{0.82} &0.238&\textbf{11.8} & \textbf{0.92} & \textbf{1.73 $|$ 31.30} & \textbf{22.97} & \textbf{20.26} & \textcolor{blue}{+159.0\%}\\ \cmidrule(lr){2-13}
        &QuIP\# & 2 &0.245 &13.3 &1.23 &0.179&11.2&1.82&2.23 $|$ 40.39 & 23.15 & 19.4 &  \\ 
        &CASP\textsubscript{QuIP\#}& 2 &\textbf{0.286} & \textbf{9.8} & \textbf{1.67}&\textbf{0.254}&\textbf{9.2}&\textbf{2.47}&\textbf{2.40} $|$ \textbf{43.12}& \textbf{24.70}  & \textbf{24.2} &\textcolor{blue}{+21.9\%}   \\
    \bottomrule
    \end{tabular}
    \vspace{-5pt}
    \caption{{Comparison results of our proposed CASP and different baselines (including GPTQ, AQLM, and QuIP\#) on LLaVA-Next-Video-7B model across video-language understanding datasets. All the results are in 2-bit precision, except GPTQ (i.e., 2.2-bit). The average relative improvement of CASP over the baselines is also provided.}
    }    
    \vspace{-5pt}
    \label{tab:videobench}
\end{table*}

\renewcommand{\arraystretch}{0.8}
\setlength{\tabcolsep}{4pt} 
\begin{table}[]
    \centering
    \footnotesize
    \begin{tabular}{l|l|c|cc|c}
         \toprule
         && & C4 & WikiText2 & {\textcolor{blue}{{Avg. Rel.}}}\\ 
         &{Method}&Bit&(PPL$\downarrow$)&(PPL$\downarrow$)&{\textcolor{blue}{{Improv.}}} \\
         \midrule
        \multirow{7}{*}{\rotatebox[origin=c]{90}{\textcolor{NavyBlue}{\textbf{Llama2-7B}}}}&\cellcolor[gray]{0.9}Original & \cellcolor[gray]{0.9}16 & \cellcolor[gray]{0.9}5.12 & \cellcolor[gray]{0.9}6.63& \\ \cmidrule(lr){2-6}
        &GPTQ &2.2 & 35.01 & 43.64 \\
        &CASP\textsubscript{GPTQ} &2.2 & \textbf{26.04} & \textbf{21.88}&\textcolor{blue}{+37.7\%} \\ \cmidrule(lr){2-6}
        &AQLM &2 & 12.57 & 9.74& \\
        &CASP\textsubscript{AQLM} &2 & \textbf{11.32} &  \textbf{8.52} &  \textcolor{blue}{+11.2\%} \\ \cmidrule(lr){2-6}
        &QuIP\# &2 & 11.00 & 8.22& \\ 
        &CASP\textsubscript{QuIP\#}&2 & \textbf{10.54} & \textbf{8.10}& \textcolor{blue}{+2.7\%} \\
    \bottomrule
    \end{tabular}
    \label{tab:LLaMA}
    \vspace{-5pt}
    \caption{
    {Comparison results of our proposed CASP and different baselines on Llama2-7B model across C4 and WikiText2 datasets. All the results are in 2-bit precision, except GPTQ (i.e., 2.2-bit).}
    }
    \vspace{-10pt}
    \label{tbl:llama}
\end{table}

\subsection{Image-Language Understanding}
Tab. \ref{tab:image-bench} presents the {numerical} results of LLaVA1.5-7B, LLaVA1.5-13B, and LLaVA-Next-7B LMMs compressed using different baselines (i.e., GPTQ, AQLM, and QuIP\#) with and without CASP on the image{-language} benchmarks. 
The relative improvement (averaged over all the benchmarks) of CASP compared to each quantization technique is highlighted in blue in the last column. Note that except for GPTQ and CASP\textsubscript{GPTQ} (i.e., 2.2 bit), all other models use average 2-bit quantization. 

As shown in Tab. \ref{tab:image-bench}, compared to CASP\textsubscript{QuIP\#}, the relative improvements achieved by CASP\textsubscript{GPTQ} and CASP\textsubscript{AQLM} are more significant due to the large performance gap between these quantization techniques and the original model. For example, on LLaVA1.5-7B, CASP\textsubscript{GPTQ} and CASP\textsubscript{AQLM} show relative improvements of 125\% and 38\%, respectively. Although QuIP\# has a smaller margin compared to the original model, CASP\textsubscript{QuIP\#} still achieves a relative improvement of {5.7\%}. It is important to note that the theoretical upper bound for relative improvement with QuIP\# is 22.6\%, assuming the original model represents the upper limit. Therefore, the 5.7\% improvement achieved by CASP\textsubscript{QuIP\#} is still quite significant.

For LLaVA1.5-13B, the performance drop of all the baselines compared to the original model is smaller, as it has more redundant parameters compared to LLaVA1.5-7B. 
{Still, CASP\textsubscript{GPTQ} and CASP\textsubscript{AQLM} respectively show a large relative performance improvement of 40\% and 32\% over GPTQ and AQLM. CASP\textsubscript{QuIP\#} obtains an improvement of {2.7\%} while the theoretical upper bound for relative improvement with QuIP\# is only 12.0\% here.} 

LLaVA-Next-7B generally outperforms LLaVA1.5-7b and LLaVA1.5-13b by increasing the input image resolution, which provides approximately 3-4X more visual tokens compared to the LLaVA1.5 models \cite{liu2023improved}. {As also discussed in the proposed method section, this potentially suggests higher sparsity and lower compression error. Thus, compared to the other models, a higher relative improvement of 141\% and 78\% is achieved by CASP\textsubscript{GPTQ} and 
CASP\textsubscript{AQLM} on LLaVA-Next-7B. Moreover, CASP\textsubscript{QuIP\#} obtains an average relative improvement of {2.3\%}, while the theoretical upper bound with QuIP\# is only 16.9\%.}

\subsection{Video-Language Understanding}
{The comparison results of the baselines and our method on LLaVA-Next-Video-7B and video-language benchmarks are presented in Tab. \ref{tab:videobench}.} 
As video-language tasks are generally more challenging, a larger performance drop after introducing low-bit compression is expected. On the other hand, {since a higher ratio of visual tokens is generated for videos, the sparsity of the attention scores increases, which results in lower attention compression error (see Section \ref{ssec:attention}).} This feature can be seen in the results in Tab. \ref{tab:videobench}, especially for CASP\textsubscript{AQLM} and CASP\textsubscript{QuIP\#} with a substantial relative  improvement of {159\% and 21\%}.

\setlength{\tabcolsep}{4pt} 
\begin{table}[]
    \centering
    \footnotesize
    \begin{tabular}{cc|ccc|c}
          \toprule
        ${Q},{K}$ & {Bit}&{{LiveB}}&{{LWilder}}&{{LCOCO}} & \textcolor{blue}{Avg.} \\ 
        {Compression}&Alloc&(PPL$\downarrow$)&(PPL$\downarrow$)&(PPL$\downarrow$)&\textcolor{blue}{(PPL$\downarrow$)}
        \\
        \midrule
        & &  38.08&138.68&7.53&\textcolor{blue}{61.43} \\ 
        & \checkmark&   27.54 &  112.91 & 7.41&\textcolor{blue}{49.28}  \\
        \checkmark& & 9.10& \textbf{10.26}& 5.63 & \textcolor{blue}{8.33}\\ 
        \checkmark& \checkmark & \textbf{8.18}&{10.46}&\textbf{5.57} & \textcolor{blue}{\textbf{8.07}}\\
    \bottomrule
    \end{tabular}
    \vspace{-5pt}
    \caption{Effect of different components on CASP\textsubscript{GPTQ} performance on LLaVA1.5-7B.}
    \vspace{-15pt}
    \label{tab:ablate}
\end{table}

\subsection{Language-Only Tasks}
To further validate the generality of the proposed CASP, we also perform experiments on Llama2-7B\footnote{https://huggingface.co/meta-llama/Llama-2-7b-hf} and two language datasets including C4 \cite{2019t5} and WikiText2 \cite{wikitext2}. {Although the motivation of sparse attention scores behind our method is more present in LMMs, the phenomenon is also observed for LLMs, which is illustrated in Fig. \ref{fig:attention}. As summarized by the results in Tab. \ref{tbl:llama}, the superiority of CASP with a relative improvement of 37\%, 11\%, and 2.7\% over GPTQ, AQLM, and QuIP\#, respectively, can be seen.}

\subsection{Ablations}

\noindent{\textbf{Components of CASP. }} Tab. \ref{tab:ablate} summarizes the effect of the two main components of CASP including $Q, K$ low-rank factorization and optimal bit allocation. 
As shown in the table, the compressed LLaVA1.5-7B has a very poor performance when none of the above-mentioned components is applied (i.e., equal to the GPTQ baseline with 2.2 bit). {Bit allocation in isolation is somewhat effective. 
However, the main improvements come from compressing attention weights. }Incorporating the low-rank compression of $Q,K$ followed the quantization of random layers to higher bits (3rd row, Tab. \ref{tab:ablate})  significantly improves the performance. {Moreover, applying the optimal bit allocation along with $Q, K$ compression (4th row, Tab. \ref{tab:ablate}) provides even more improvement. We argue that the proposed bit allocation has limitations to address in future work including determining the optimal bits for each weight instead of each layer. }

\begin{figure}[]
    \centering
    \includegraphics[width=0.45\textwidth]{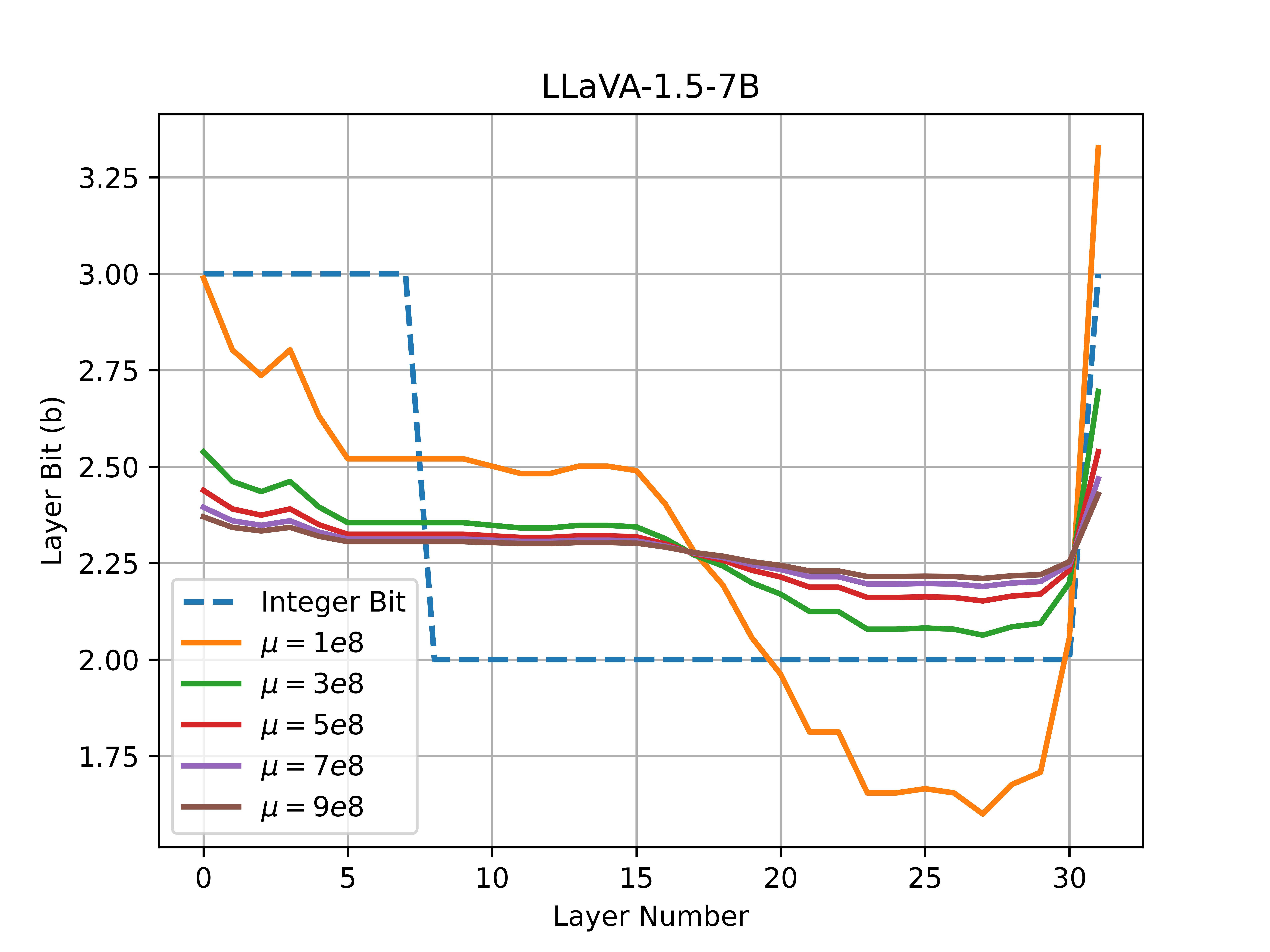}
    \vspace{-8pt}
    \caption{{Optimal bit computation by Eq. \ref{eq:quant2} for different $\mu$ values. Note that the layer bit (i.e. $b_l$ in Eq. \ref{eq:quant2}) only accounts for the compression obtained from quantization, not the low-rank decomposition. Therefore, the average layer bit from the above plot is not the actual average bit of the model (i.e. $B_{\text{avg}}$ in Eq. \ref{eq:quant2}).}}
    \vspace{-3pt}
    \label{fig:optimal-bit}
\end{figure}

\noindent\textbf{Optimal Bit Allocation.} 
Fig. \ref{fig:optimal-bit} demonstrates the returned optimal bit by Eq. \ref{eq:quant2} for different values for the regularization parameter $\mu$. This study was done on LLaVA1.5-7B model. The non-integer optimal bit can be allocated by changing the group size for GPTQ and by changing the codebook size and number of codebooks for AQLM and QuIP\#. For simplicity, we convert the optimal bit to integer values (dashed lines in Fig. \ref{fig:optimal-bit}) and use the integer bit in our experiments. Generally, we observe that the first layers and the last layer are allocated higher bits that show the importance of these layers in the model.   

\noindent\textbf{Calibration Dataset.} {In this study, we analyze the impact of using multi-modal vs. text-only calibration datasets on the performance of the CASP\textsubscript{AQLM} with LLaVA1.5-7B. Specifically, we use LLaVA-Instruct-150K as a multi-modal dataset \cite{liu2023llava}, and RedPajama \cite{together2023redpajama} and C4 \cite{2019t5} as text-only datasets (1024 samples from each). The corresponding results in Tab. \ref{tab:calib-data} indicate that optimizing the quantization procedure with text-only calibration datasets provides lower PPL on average. We argue that since the underlying LLM has initially been pre-trained with a significant amount of language-only data, it is more aligned with similar types of calibration data for quantization.} 

\setlength{\tabcolsep}{4pt} 
\begin{table}[]
    \centering
    \footnotesize
    \begin{tabular}{c|ccc|c}
          \toprule
         Calibration&{LiveB}&{LWilder}&{LCOCO} & \textcolor{blue}{Avg.} \\ 
          Dataset&(PPL$\downarrow$)&(PPL$\downarrow$)&(PPL$\downarrow$)& \textcolor{blue}{(PPL$\downarrow$)} \\
         \midrule
        RedPajama & 7.86  & \textbf{8.24} & \textbf{5.70} & \textcolor{blue}{7.26}\\ 
        C4 & \textbf{6.82} & 8.96 & 5.70 & \textcolor{blue}{7.16}\\
        LLaVA-Instruct-150K & 7.84 &13.68&5.72&\textcolor{blue}{{9.08}}\\ 
    \bottomrule
    \end{tabular}
    \vspace{-5pt}
    \caption{CASP\textsubscript{AQLM} performance with different calibration datasets on LLaVA1.5-7B.}
    \vspace{-5pt}
    \label{tab:calib-data}
\end{table}
\setlength{\tabcolsep}{2pt} 
\begin{table}[]
    \centering
    \footnotesize
    \begin{tabular}{l|l|c|ccc|c}
          \toprule
          \small
         &&&{LiveB}&{LWilder}&{LCOCO} & \small{\textcolor{blue}{{Avg. Rel.}}}  
         \\  
         &{Method}&Bit&(PPL$\downarrow$)&(PPL$\downarrow$)&(PPL$\downarrow$)&\small{\textcolor{blue}{{Improv.}}}
         \\
         \cmidrule(lr){1-7}
        \multirow{7}{*}{\rotatebox[origin=c]{90}{\textcolor{NavyBlue}{\textbf{LLaVA1.5-7B}}}}&\cellcolor[gray]{0.9}Original  & \cellcolor[gray]{0.9}16 & \cellcolor[gray]{0.9}5.45 & \cellcolor[gray]{0.9}4.24 & \cellcolor[gray]{0.9}4.51 \\  \cmidrule(lr){2-7}
        &GPTQ & 3 & 9.07 & 6.28 & 5.03 \\
        &CASP\textsubscript{GPTQ} & 3& \textbf{6.07}&\textbf{4.85}&\textbf{4.80} & \textcolor{blue}{+20.1\%}\\  \cmidrule(lr){2-7}
        &AQLM & 3&7.64 &4.54&4.72\\
        &CASP\textsubscript{AQLM} & 3& \textbf{6.06} & 4.88 &4.86 & \textcolor{blue}{+3.4\%}\\  \cmidrule(lr){2-7}
        &QuIP\# & 3 & \textbf{5.68} & 4.53 & \textbf{4.62} \\ 
        &CASP\textsubscript{QuIP\#}& 3 & 5.69 & \textbf{4.51} & 4.63 & \textcolor{blue}{+0.01\%}  \\
    \bottomrule
    \end{tabular}
    \vspace{-5pt}
    \caption{Experiments with 3-bit compression. 
    }
    \vspace{-5pt}
    \label{tab:3bit}
\end{table}

\setlength{\tabcolsep}{0.6pt} 
\begin{table}[]
    \centering
    \footnotesize
    \begin{tabular}{c|cccc}
        \toprule
         {Attention}& {Llama2} & {LLaVA1.5} & {LLaVA1.6} & {LLaVA-Next-Video} \\ 
          {Compression}& (C4)& (LWilder)&(LWilder) &(VideoChatGPT)\\ 
          {Ratio}& \#VT=0& \#VT=575 & \#VT$\approx$2500& \#VT=4600 \\ \midrule \rowcolor[gray]{0.9}
        Original & 7.28 & 4.24 & 3.77 &7.09 \\ 
        50\% & 7.57 \textcolor{blue}{($\downarrow$ 4\%)} & 4.38 \textcolor{blue}{($\downarrow$ 3\%)} & 3.88 \textcolor{blue}{($\downarrow$ 2\%)} & 7.14 \textcolor{blue}{($\downarrow$ 0.7\%)} \\
        75\% & 8.20 \textcolor{blue}{($\downarrow$ 12\%)} &4.72 \textcolor{blue}{( $\downarrow$ 11\%)}& 4.13 \textcolor{blue}{( $\downarrow$ 9\%)} &7.40 \textcolor{blue}{( $\downarrow$ 4\%)} \\ \bottomrule
    \end{tabular}
    \vspace{-5pt}
    \caption{Experiments on the effect of number of visual tokens in different models LLM and LMMs in terms of PPL. 
    }
    \vspace{-10pt}
    \label{tab:atten-compression}
\end{table}

\noindent\textbf{Higher Quantization Bits.} 
The comparison results of CASP and the baseline quantization techniques with 3-bit precision are summarized in Tab. \ref{tab:3bit}. It is shown that the performance drop of the baselines compared to the original model is marginal when it comes to 3-bit. Similarly, the relative improvements of CASP is smaller since the theoretical upper bound for relative improvements is only 7\% for QuIP\#. {Compared to QuIP\#, CASP has no improvement as both have almost reached the original model performance.}

\noindent\textbf{Effect of Vision Tokens Ratio.} Tab. \ref{tab:atten-compression} shows the experimental results for different models with various number of vision tokens (\#VT).
{In this section,} we only apply the first phase of CASP, that is low-rank decomposition of $W_q$ and $W_k$ and removing 50\% and 75\% of the eigenvalues. 
{The results in the table show that as the number of vision tokens (sparsity of the attention map) increases, less performance drop after compression is observed. For example, Llama2 has the highest drop with 0 visual tokens, while LLaVA-Next-Video with 4.6K visual tokens has the lowest performance drop in both 50\% and 75\% compression ratios.}

\noindent\textbf{More Analysis and Results.} 
{
The computational complexity analysis, the effect of calibration dataset size, further numerical experiments on more datasets, and qualitative results are given in the supplementary materials.}

\section{Conclusion}
In this work, we proposed a low-bit model compression technique for LMMs. Our insights for CASP arise from empirical observations that the attention maps in LMMs are highly sparse. We theoretically and experimentally showed that the Query and Key weight matrices can be compressed with negligible performance drop in the model. We also proposed an optimal bit allocation approach to obtain an average target bit outperformed state-of-the-art low-bit model compression techniques. Our extensive experimental results on different LMMs and benchmarks for image- and vide-language understanding showed the effectiveness and generality of the proposed method. 
The insight from this work has a broader impact on the architecture design of LMMs resulting in more efficient attention mechanism.  

\clearpage

{
    \small
    \bibliographystyle{ieeenat_fullname}
    \bibliography{main}
}

\clearpage
\setcounter{page}{1}
\maketitlesupplementary
{This supplementary material includes the computational complexity analysis, further numerical experiments, an ablation study on the calibration dataset size, and qualitative results.} We also discuss the limitations and broader impact of this work.

\section{Computational Complexity}

In this section, we present the computational complexity analysis of CASP compared to the baselines. Tab. \ref{tab:runtime} shows the results on LLaVA-Next-Video-7B (8 frames) \cite{zhang2024llavanextvideo} with "Eager" attention, a batch size of 1, and a maximum/minimum new token count of 128. We provide the prefilling time in seconds and throughput in tokens per second (Tok/s). Additionally, we report the prefilling peak memory, end-to-end peak memory, and model size.

Note that the quantization procedure {generally} involves two criteria that can affect the inference time: 1) Matrix multiplication of low-precision tensors, which is often faster than float tensors. 2) Dequantization at the inference stage to FP16, which introduces overhead. 
Tab. \ref{tab:runtime} shows the inference time of AQLM \cite{aqlm} and QuIP\# \cite{quipsharp} compared with the original model in FP16. 
{Comparing QuIP\# and AQLM, QuIP\# is faster via fusing query, key, and value weight matrices in the attention layer and fusing gate and up weight matrices in the MLP layer. }

CASP contains two components that impact the inference time: 1) Low-rank factorization of $W_q$ and $W_k$. 2) Quantizing important layers to higher bits (e.g., 3-bit).
Compressing $W_q$ and $W_k$ via low-rank decomposition (i.e., removing a high percentage of eigenvalues from the Q and K weights) directly reduces FLOPs, making inference faster. {In other words, regardless of the hardware and kernel design, low-rank factorization always provides run-time improvement as most of the parameters are removed.} As the second row of Tab. \ref{tab:runtime} shows, CASP\textsubscript{Original}, i.e., the FP16 model with 75\% compression of $W_q$ and $W_k$, results in nearly 4\% speed-up due to smaller weight matrices. 

On the other hand, quantizing important layers to higher bits may introduce overhead compared to uniformly quantizing all layers to 2-bit. This is because 3-bit quantized models are slightly slower than the 2-bit ones \cite{aqlm,quipsharp}. Overall, CASP does not introduce any overhead for the baselines. In some cases such as CASP\textsubscript{AQLM}, it can slightly improve the inference speed {due to the low-rank factorization}.
It should be noted that our primary goal in this work is not to achieve faster inference over the baselines but to enhance their performance with the same model size, memory, and inference time. 

{Tab. \ref{tab:runtime} also compares the prefilling and end-to-end peak memory of CASP with the baselines. For a fair comparison, we matched the model size of CASP with the baselines, ensuring all 2-bit quantized checkpoints are 2.7GB. CASP’s peak memory is slightly higher than the baseline due to optimal bit allocation. This peak memory is influenced by the higher bits allocated to important layers and the extent of low-rank compression applied to $W_q$ and $W_k$.}

\setlength{\tabcolsep}{1.6pt} 
\begin{table}
    \centering
    \footnotesize
    \begin{tabular}{l|c|cc|cc|c}
    \toprule
        Method & Bit& Prefill  &   Throughput& Prefill& End-to-End & Model \\ 
        & &Time (s) &  (Tok/s) &Peak-Mem  &  Peak-Mem & Size \\ 
        & & &  & (GB) & (GB) & (GB)  \\ \midrule \rowcolor[gray]{0.9}
        Original &16& 0.41  & 2.2 &13.5& 13.6 & 13.5\\ \rowcolor[gray]{0.9}
        CASP\textsubscript{Original}& 16&0.39 & 2.3 & 12.0  & 12.1&13\\ \midrule 
        AQLM & 2&0.51 & 1.8 & 3.2 &3.4 &2.7 \\
        CASP\textsubscript{AQLM}&2& 0.50 & 1.9 & 3.1 & 3.3&2.7\\ \midrule
        QuIP\#&2& 0.39 & 2.3 & 3.2 &3.4&2.7\\
        CASP\textsubscript{QuIP\#}&2& 0.39 & 2.3 & 3.4 & 3.6&2.7\\
        \bottomrule
    \end{tabular}
    \caption{Runtime and memory usage of the baselines and CASP. CASP does not introduce any overhead compared to the baselines.}
    \label{tab:runtime}
    
\end{table}

\setlength{\tabcolsep}{2.5pt} 
\begin{table}[]
    \centering
    \footnotesize
    \begin{tabular}{l|c|cccc|c}
          \multicolumn{7}{c}{\textcolor{NavyBlue}{\textbf{LLaVA-1.5-7B}}} \\
          \toprule
         & &{NoCaps}&{COCO17}&{Flick30K}&{GQA} & \textcolor{blue}{Avg.}  \\
         & Bit&(CIDEr$\uparrow$)&(CIDEr$\uparrow$)&(CIDEr$\uparrow$)&(EM $\uparrow$) & \textcolor{blue}{Rel Imp.}\\ \midrule \rowcolor[gray]{0.9}
        Original  & 16 & 0.102  & 0.106 & 0.74 & 0.61 \\ \midrule
        GPTQ & 2.2 & 0.53 & 0.62 & 0.38 & 0.13  \\
        CASP\textsubscript{GPTQ}  & 2.2 & \textbf{0.92} & \textbf{0.100} & \textbf{0.64} & \textbf{0.52}&\textcolor{blue}{+125\%} \\ \midrule
        AQLM & 2 & 0.73 & 0.87 & 0.57 & 0.43  \\
        CASP\textsubscript{AQLM}  & 2& \textbf{0.91} & \textbf{0.107} & \textbf{0.68} & \textbf{0.53} &\textcolor{blue}{+22\%} \\ \midrule
        QUIP\# & 2 & {0.102} &0.103&0.75&0.57\\ 
        CASP\textsubscript{QuIP\#} & 2 & \textbf{0.102}&0.102 &\textbf{0.77}&\textbf{0.57}&\textcolor{blue}{+0.5\%}\\
    \bottomrule
    \end{tabular}
    \caption{Further quantitative results on open-ended QA tasks and GQA dataset with LLaVA-1.5-7B.}
    \label{tab:open-qa}
\end{table}

\setlength{\tabcolsep}{4pt} 
\begin{table}[]
    \centering
    \footnotesize
    \begin{tabular}{l|ccc|c}
          \toprule
        Calibration&{{LiveB}}&{{LWilder}}&{{LCOCO}} & \textcolor{blue}{Avg.} \\ 
        {Size}&(PPL$\downarrow$)&(PPL$\downarrow$)&(PPL$\downarrow$)&\textcolor{blue}{(PPL$\downarrow$)}
        \\
        \midrule
        128& 7.8 & 9.0 & 5.9 & \textcolor{blue}{7.5}  \\
        256& 7.8& 8.5& 5.8 & \textcolor{blue}{7.3}\\ 
        512& 7.9& 8.3& 5.7 & \textcolor{blue}{7.3}\\ 
        1024& 7.9 & 8.2 & 5.7 & \textcolor{blue}{7.2} \\
    \bottomrule
    \end{tabular}
    \caption{Experiment on the calibration data size using CASP\textsubscript{AQLM} with LLaVA-1.5-7B.} 
    \vspace{-15pt}
    \label{tab:calib-size}
\end{table}

\setlength{\tabcolsep}{3pt} 
\renewcommand{\arraystretch}{1}
\begin{table*}[tbh!]
\footnotesize
{%
\begin{tabular}{l|p{3cm}lll}
\toprule
& Dataset               & Task              & Metric & System Prompt \\ \midrule \multirow{13}{*}{\rotatebox[origin=l]{90}{\textcolor{black}{\textbf{Image-Language }}}} 
&COCO-2017~\cite{lin2015microsoft}             & Image Captioning  & CIDEr                                                                 & Provide a one-sentence caption for the provided image.                                                                                                                     \\ \cline{2-5}
&Flicker30k~\cite{young-etal-2014-image}           & Image Captioning  & CIDEr                                                                 & Provide a one-sentence caption for the provided image.                                                                                                                     \\\cline{2-5}
&GQA~\cite{gqa}                   & CE-VQA            & Eaxct Match                                                           & Answer the question using a single word or phrase.                                                                                                                         \\\cline{2-5}
&MMBench~\cite{mmbnech}               & MC-VQA            & Accuracy                                                              & Answer with the option's letter from the given choices   directly.                                                                                                         \\\cline{2-5}
&MME~\cite{mme}                   & CE-VQA            & Perception Score                                                      & Answer the question using a single word or phrase.                                                \\\cline{2-5}
&LiveBench~\cite{livebench}   & OE-VQA  & PPL & N/A  \\\cline{2-5}
&LLaVA-Bench-Wilder~\cite{Jelinek1977PerplexityaMO}   & OE-VQA  & PPL & N/A  \\\cline{2-5}
&LLaVA-Bench-COCO~\cite{Jelinek1977PerplexityaMO}   &  Image Captioning & PPL & N/A  \\\cline{2-5}
&MMU~\cite{mmu}                   & CE-VQA,OE-VQA & Accuracy                                                             & \begin{tabular}[c]{@{}l@{}}Answer with the option's letter from the given choices   directly, OR \\ Answer the question using a single word or phrase.\end{tabular}        \\\cline{2-5}
&Nocaps~\cite{Agrawal_2019}                & Image Captioning  & CIDEr                                                                 & Provide a one-sentence caption for the provided image                                                                                                                      \\\cline{2-5}
&ScienceQA-Image~\cite{scieniceqa}       & Visual reasoning  & Exact Match                                                          & Answer with the option's letter from the given choices   directly.                                                                                                         \\\cline{2-5}
&SeedBench-Image~\cite{seedbench}        & MC-VQA            & Accuracy                                                              & Answer with the option's letter from the given choices   directly.                                                                                                         \\ \cline{2-5}

\midrule \midrule \multirow{4}{*}{\rotatebox[origin=c]{90}{\textcolor{black}{\textbf{~~Video-Language}}}} 
&ActivityNet~\cite{activitynetqa}           & CE-VQA            & \begin{tabular}[c]{@{}c@{}}Accuracy/\\ GPT-Assisted score\end{tabular} & Answer the question using a single word or phrase.                                                                                                                \\\cline{2-5}
&VideoChatGPT-temporal~\cite{videochatgptT} & OE-VQA            & 
\begin{tabular}[c]{@{}c@{}}Rouge, PPL, and\\ GPT-Assisted scores\end{tabular}
& Evaluate the temporal accuracy of the prediction compared to  the answer.$^*$                                                
                    \\\cline{2-5}
&VideoDetailCaption~\cite{Jelinek1977PerplexityaMO} & OE-VQA & \begin{tabular}[c]{@{}c@{}}Rouge, PPL, and\\ GPT-Assisted scores\end{tabular} & N/A \\\cline{2-5}
&VideoMME (VMME)~\cite{fu2024video} & MC-VQA & Accuracy & Answer with the option's letter from the given choices directly. \\\cline{2-5}
&NextQA~\cite{nextqa}                & CE-VQA            & WUPS                                                                  & Answer a question using a short phrase or sentence.  \\
\bottomrule
\end{tabular}
}
\caption{Details of the datasets, the corresponding tasks, metrics, and prompts used in our experiments. CE-VQA: Closed-Ended Visual Question Answering, OE-VQA: Open-Ended Visual Question Answering, MC-VQA: Multiple-Choice Visual Question Answering. $^*$: Only the main sentence from the prompt is shown here.}
\label{tbl:dataset_detail}
\end{table*}

\section{Further Quantitative Results}
\label{sec:extra_ds}
{In the main manuscript, the experimental results on 5 multi-choice QA datasets for image-language understanding were reported.}
In this section, Tab. \ref{tab:open-qa} presents additional quantitative results on image captioning datasets such as NoCaps \cite{Agrawal_2019}, COCO-Caption \cite{lin2015microsoft}, and Flickr30K \cite{young-etal-2014-image}, as well as GQA \cite{gqa}. The primary evaluation metric used for open QA and image captioning tasks is CIDEr (Consensus-based Image Description Evaluation) \cite{cider}, which measures the similarity between a generated caption and a set of reference captions. As summarized in Tab. \ref{tab:open-qa}, CASP obtains 125\% and 22\% average relative improvements over GPTQ and AQLM. QUIP\#  almost obtains the same results as the FP16 model and even outperforms the FP16 model in the Flickr30K dataset. However, we still observe 0.5\% average relative improvements with CASP\textsubscript{QuIP\#}.  

\section{Calibration Dataset Size}
Tab. \ref{tab:calib-size} demonstrates experiments on the number of samples in the calibration dataset used for CASP\textsubscript{AQLM} with LLaVA-1.5-7B. We observe slight performance improvements with increasing the calibration size from 128 samples to 1024 samples. 
Although increasing the size of the calibration dataset improves the overall performance of the model, it also increases the cost and time of the calibration and optimization procedure for quantization and low-rank factorization.

\section{CASP and KV Cache Quantization}
{KV cache compression has emerged as a critical technique to optimize memory efficiency in large language models by reducing the size of the key-value cache used during inference. One recent method for KV cache quantization is KIVI \cite{liu2024kivi}, which achieves significant reductions in storage requirements while preserving model performance.
On the other hand, CASP focuses on weight-only compression, targeting the model's parameters to achieve similar efficiency gains. These two approaches are orthogonal, meaning they operate on different components of the model and can be combined to further enhance overall compression.}

{As KIVI and CASP are orthogonal methods, we have combined them. Tab. \ref{tab:kvcache} demonstrates the results on TruthfulQA (BLEU Score$\uparrow$) using Llama2-7B as the base model. KV cache is quantized to 2 bits and model weights are quantized to 2.2 bits (on average). As seen, CASP\textsubscript{GPTQ}+KIVI offers a significant improvement over GPTQ+KIVI.}
\setlength{\tabcolsep}{2pt} 
\begin{table}[h!]
    \vspace{-8pt}
    \centering
    \footnotesize
    \begin{tabular}{cc|cc|cc} \toprule 
        Base & Base+KIVI &GPTQ&GPTQ+KIVI 
        &CASP\textsubscript{GPTQ}&CASP\textsubscript{GPTQ}+KIVI   \\ \midrule
         26.0& 21.6&5.0&2.8&\textbf{23.5}&\textbf{11.4}\\ \bottomrule
    \end{tabular}
    \caption{CASP combined with KV cache quantization.}
    \vspace{-15pt}
    \label{tab:kvcache}
\end{table}
\section{CASP vs. Low-Rank Decomposition}
{Applying simple low-rank decomposition to ALL weight matrices results in significantly worse performance than CASP. This is because only $W_q$ and $W_k$ are low-rank in LMMs and LLMs. Tab. \ref{tab:lora-casp} shows the results of CASP with SOTA low-rank decomposition methods SVD-LLM \cite{WangX24} and MoDeGPT \cite{LinC24} under extreme compression. We use LLama2-7B as the base model and report perplexity (PPL$\downarrow$) on the Wikitext dataset.}

\setlength{\tabcolsep}{3pt} 
\begin{table}[h!]
\vspace{-8pt}
\footnotesize
    \centering
    \begin{tabular}{c|c|c|c|c}
    \toprule 
       \multicolumn{2}{c|}{80\% Compression ({$\approx$3.2 bits})}&\multicolumn{3}{c}{87\% Compression ({3 bits})} \\  \hline
         SVD-LLM & ModeGPT & CASP\textsubscript{GPTQ}&CASP\textsubscript{AQLM}&CASP\textsubscript{QuIP\#} \\
        276.4&245.8&21.8&8.5&8.1\\ \bottomrule
    \end{tabular}
    \caption{CASP vs. low-rank decomposition methods.}
    \label{tab:lora-casp}
    \vspace{-12pt}
\end{table}

\section{Further Analysis on Bit Allocation}
{The optimal bit allocations returned by our method are typically non-integer. To ensure simplicity and compatibility across various quantization techniques, we rounded these values to integers. Calculating exact non-integer average bits for each layer would require modifying the codebook to accommodate non-predefined values for techniques such as AQLM and QuIP\#. This adjustment, however, would necessitate the creation of new GPU kernels for decoding during inference—one kernel for each layer. While using non-integer bits could potentially yield better results, exploring this avenue is left as future work.}

{In our experiments, we computed the optimal bit allocation for each individual layer in the model. However, since adjacent layers often share similar levels of importance, we investigated the possibility of sharing bit allocations across adjacent layers. Specifically, we tested shared optimal bit allocations for every three layers on LLaVA-1.5-7B. This approach resulted in only a negligible reduction of 0.7 seconds in overall computation time, which is insignificant compared to the total quantization times: 40 minutes for GPTQ, 2 hours for QuIP\#, and 6 hours for AQLM.}

\section{Datasets, Tasks, and Metrics}
We briefly introduced the 8 image-language and 5 video-language datasets used in the experiments of the main manuscript. In addition, the system prompt (instruction) used to get output results for each dataset was given. 
{Similar to the experiments on LLMs, when measuring perplexity we do not provide any system prompt \cite{FrantarE23}.}
The details of datasets used for image-language and video-language understanding tasks are presented in Tab.~\ref{tbl:dataset_detail}, which also includes the extra 4 datasets discussed in Section \ref{sec:extra_ds}. 

As shown in the table, diverse range of tasks including image captioning, visual reasoning, open-ended visual question answering, closed-ended visual question answering, and multiple-choice visual question answering are used to evaluate the performance of the baseline methods compared with ours. Note that the system prompts are the default prompts provided in the lmms-evals evaluation package \cite{zhang2024lmmsevalrealitycheckevaluation}.

\section{Qualitative Results}
In this section, we provide qualitative results from LiveBench \cite{livebench}, COCO-Caption \cite{lin2015microsoft}, and LLaVA-Bench-Wilder \cite{liu2023llava} datasets.

LiveBench includes screenshots from news web pages, with multiple questions asking for details about each image. Fig. \ref{fig:qualitative1} and \ref{fig:qualitative2} show two randomly chosen examples from this dataset. Below each image, we display the responses from LLaVA-1.5-7B (FP16), baselines (GPTQ, AQLM, and QuIP\#), and CASP. Each response is scored by GPT-4o out of 10. CASP consistently improves the baseline responses by approximately {1.5 points}.

Fig. \ref{fig:qualitative3} and \ref{fig:qualitative4} present two samples from the COCO-Caption dataset, which includes images with multiple short captions for each image. This task is generally easier compared to LiveBench. We observe consistent improvements in responses by CASP, with an average increase of 2.6 points. In Fig. \ref{fig:qualitative3}, CASP\textsubscript{QuIP\#} addresses the redundancy in QuIP\#'s answer by including most of the important elements in the picture. In Fig. \ref{fig:qualitative4}, a major element, “Man hangs off the side of the motorcycle,” is overlooked by both the FP16 model and quantized models. However, CASP\textsubscript{QuIP\#} eliminates unnecessary information from the FP16 response (e.g., “A backpack can be seen…”). Comparing the responses of QuIP\# and CASP\textsubscript{QuIP\#}, the latter adds important aspects such as “the motorcycle is leaning over” and “the rider is leaning into the turn.”

Fig. \ref{fig:qualitative5} and \ref{fig:qualitative6} are from LLaVA-Bench-Wilder. The questions are complex and include “memes” that require the model to understand indirect meanings in the pictures. CASP\textsubscript{QuIP\#} scores are equal to or better than the FP16 model in these examples. Overall, these qualitative results show the effectiveness and superiority of CASP compared to the baselines in terms of basic understanding and addressing important details in the images.

\section{Limitations and Future Work}
This work has some limitations that need to be addressed in future research. 
The low-rank factorization method used in this work is not quantization-friendly, leading to more outliers in the factorized matrices compared to the original weight matrices. Addressing this issue could improve CASP’s results in future work.

We also observe that the extreme compression regime applied in CASP decreases accuracy for samples with small images and complex questions, as there is less redundancy in the attention. Providing a dynamic rank selection for such cases, similar to the dynamic visual token of LLaVA-1.6 could address this problem. In this study, we presented results without fine-tuning the quantized models. Future research should explore efficient layer-wise fine-tuning to further enhance the performance of quantized models {combined with low-rank factorization}.

\begin{figure*}[t!]
    \centering
    \includegraphics[width=0.9\linewidth]{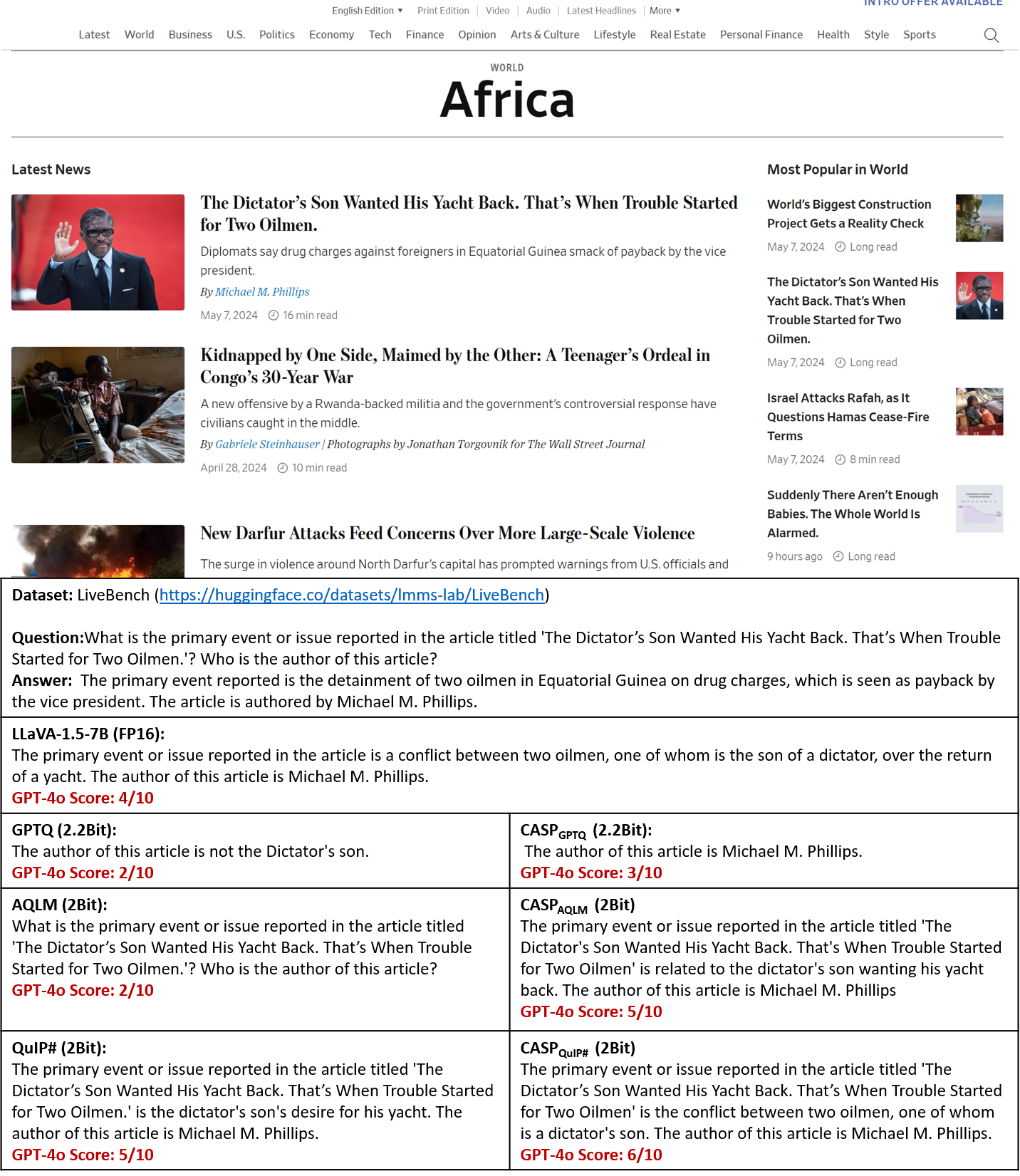}
    \caption{Qualitative results from LiveBench dataset. The GPT-4o scores out of 10 are shown for each method.
    }
    \label{fig:qualitative1}
\end{figure*}

\begin{figure*}[t!]
    \centering
    \includegraphics[width=0.9\linewidth]{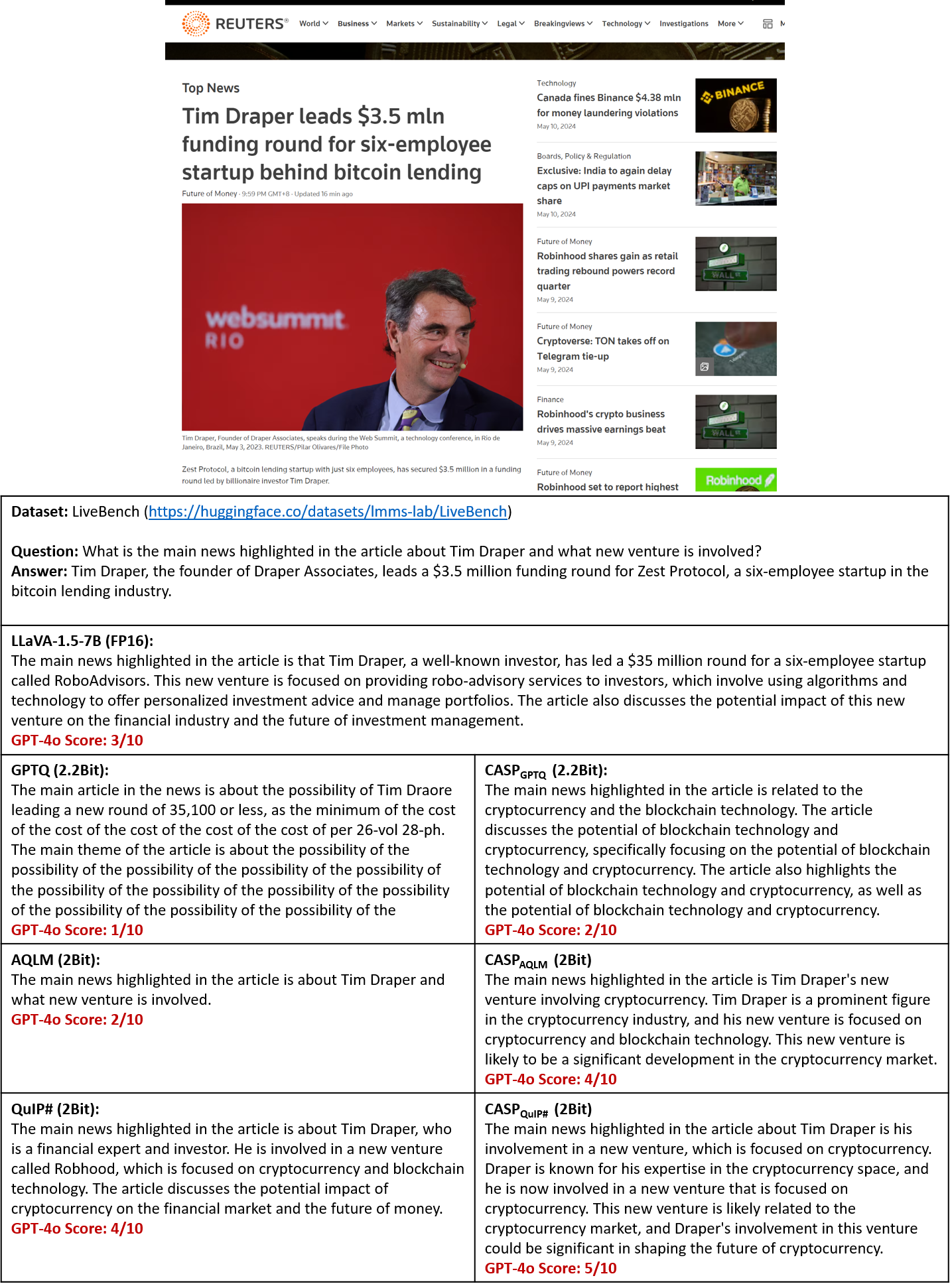}
    \caption{Qualitative results from LiveBench dataset. The GPT-4o scores out of 10 are shown for each method.
    }
    \label{fig:qualitative2}
\end{figure*}

\begin{figure*}[t!]
    \centering
    \includegraphics[width=0.9\linewidth]{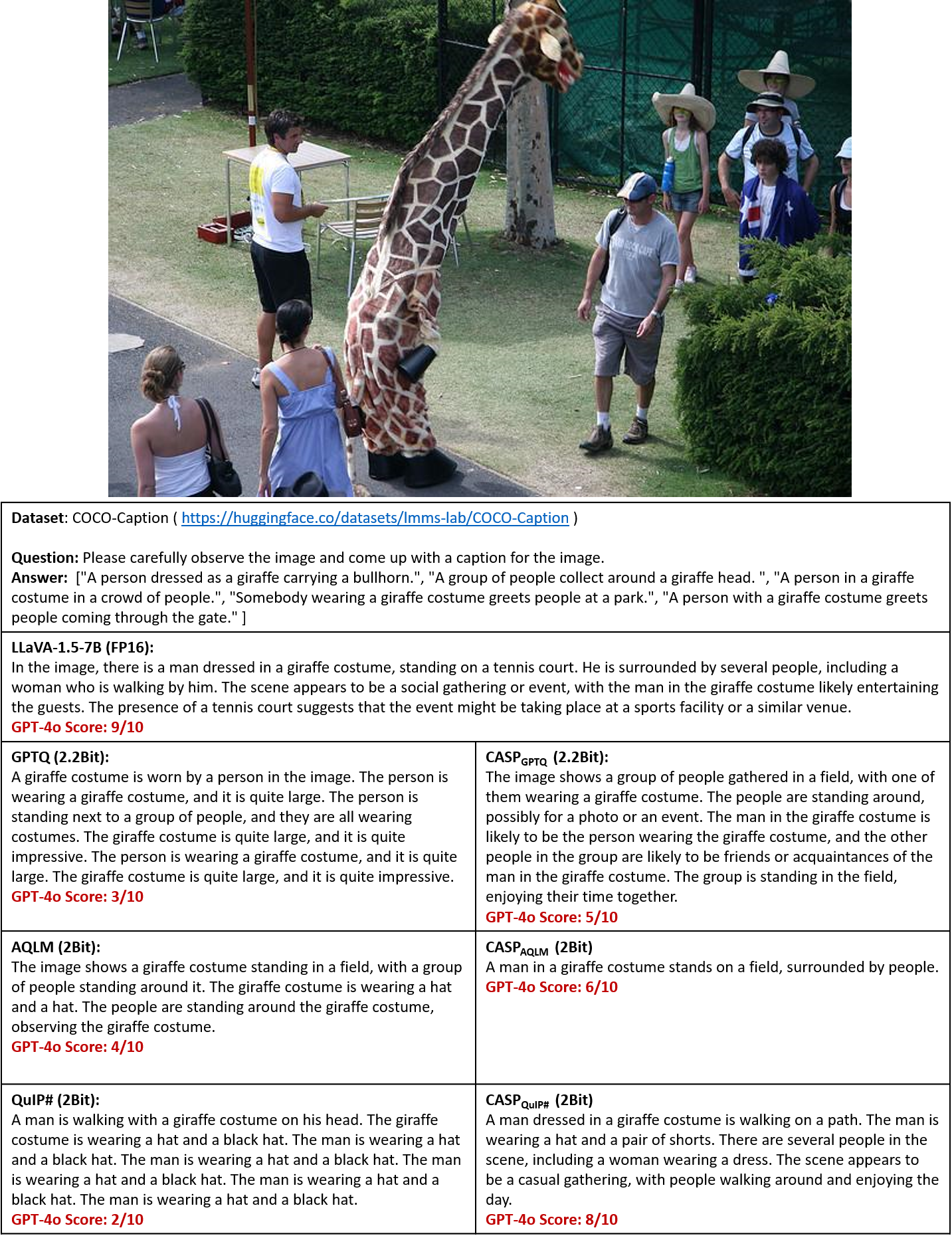}
    \caption{Qualitative results from COCO-Caption dataset. The GPT-4o scores out of 10 are shown for each method.
    }
    \label{fig:qualitative3}
\end{figure*}

\begin{figure*}[t!]
    \centering
    \includegraphics[width=0.9\linewidth]{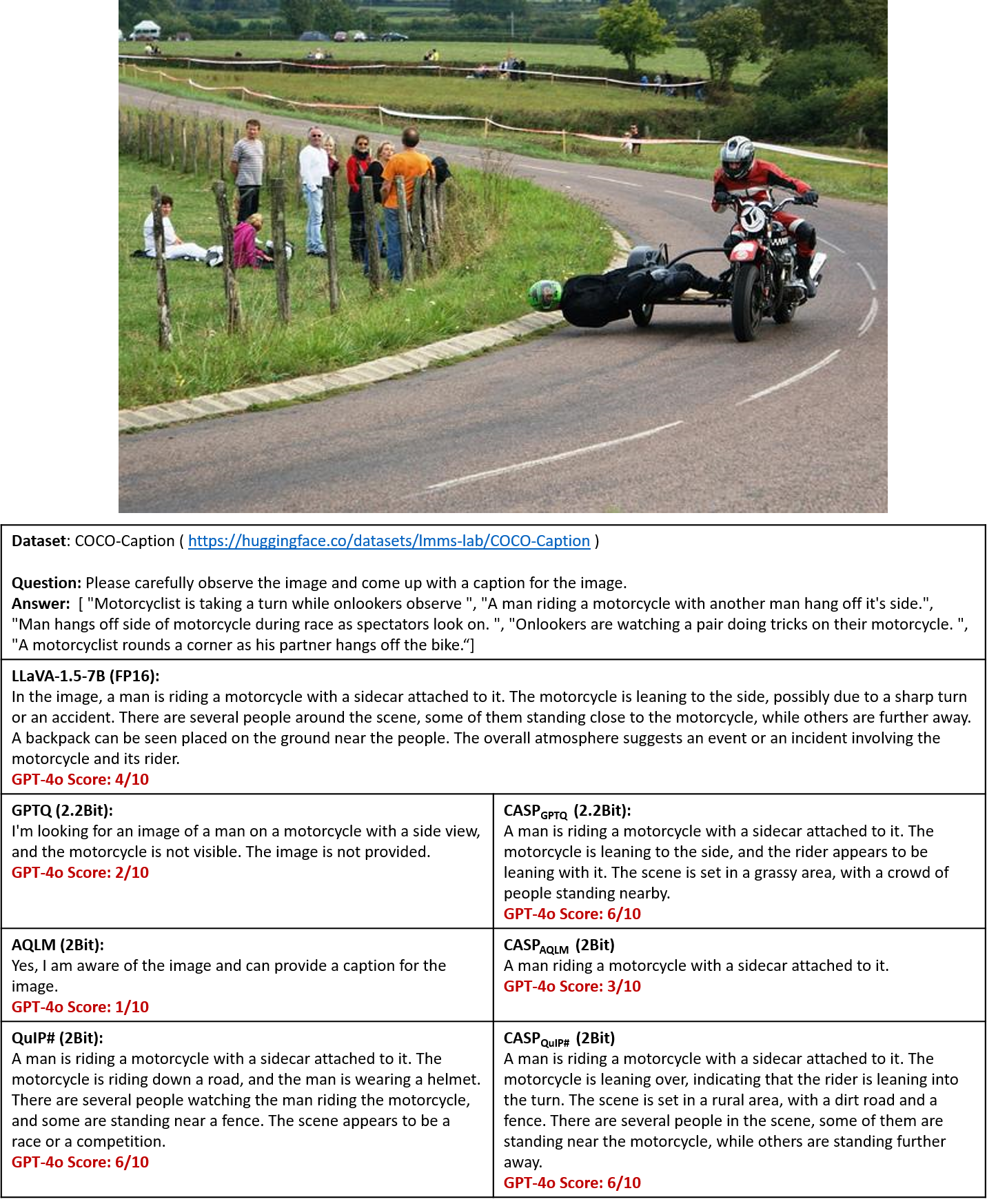}
    \caption{Qualitative results from COCO-Caption dataset. The GPT-4o scores out of 10 are shown for each method.
    }
    \label{fig:qualitative4}
\end{figure*}

\begin{figure*}[t!]
    \centering
    \includegraphics[width=0.9\linewidth]{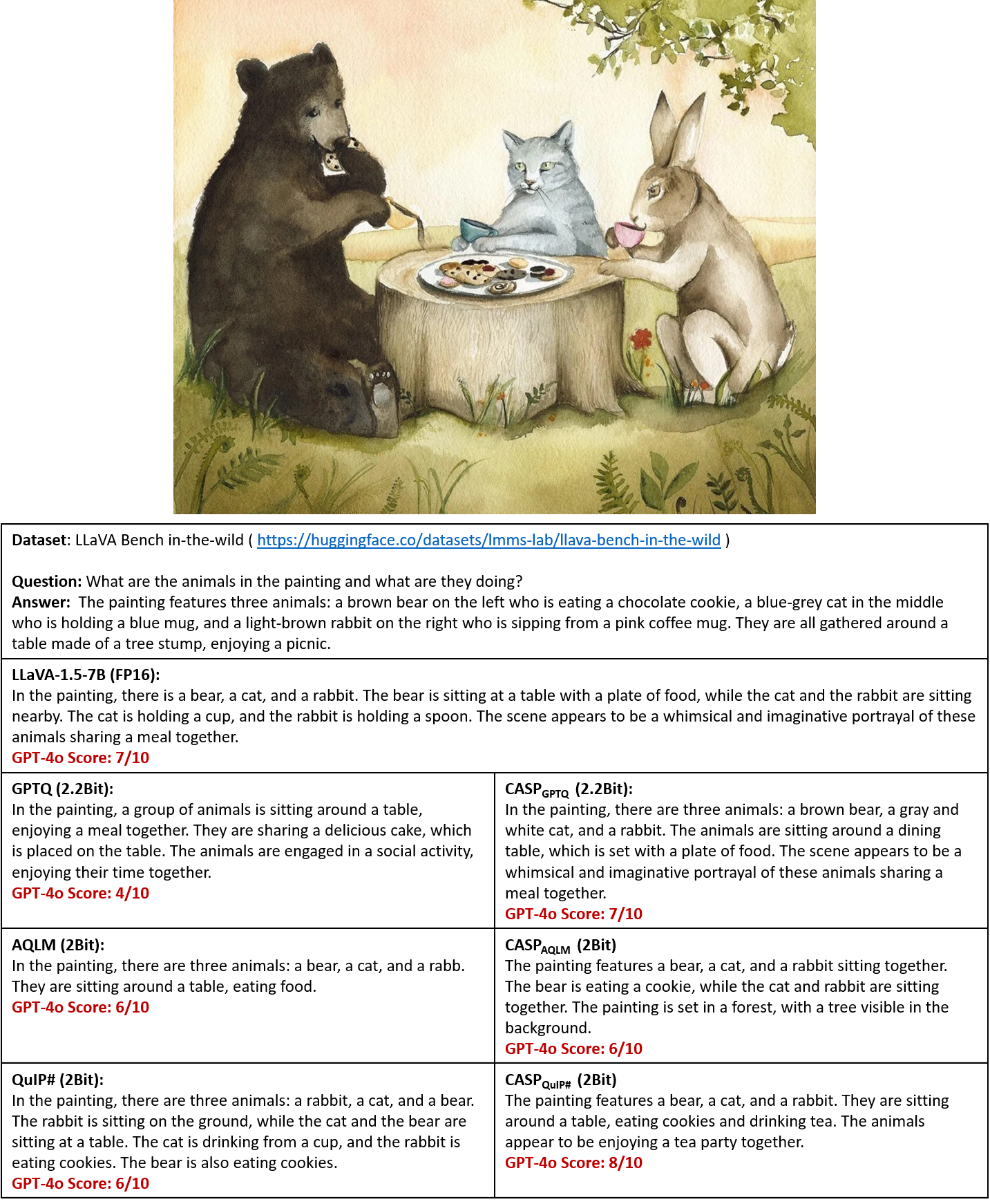}
    \caption{Qualitative results from LLaVA Bench in-the-wild dataset. The GPT-4o scores out of 10 are shown for each method.
    }
    \label{fig:qualitative5}
\end{figure*}

\begin{figure*}[t!]
    \centering
    \includegraphics[width=0.9\linewidth]{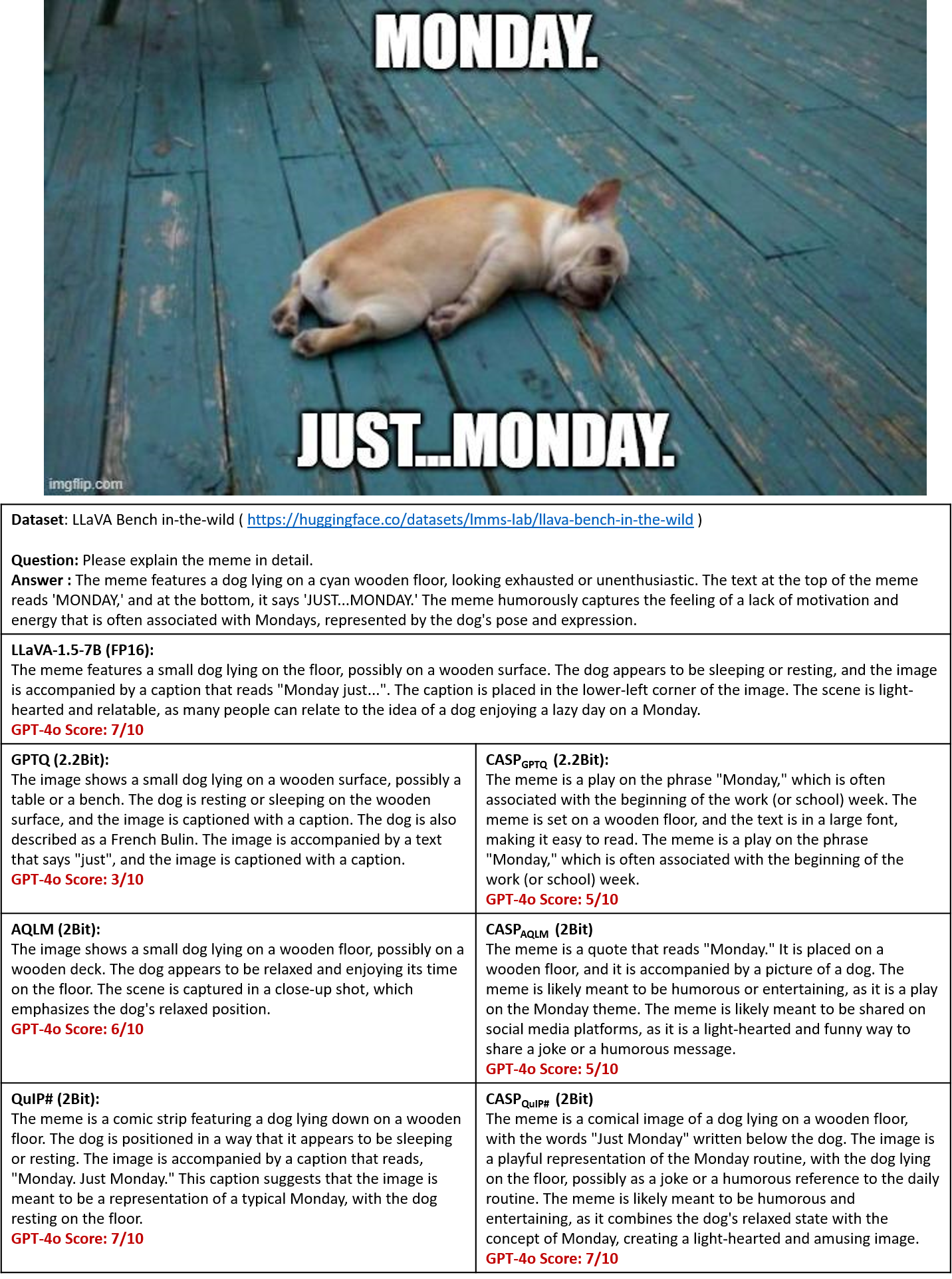}
    \caption{Qualitative results from LLaVA Bench in-the-wild dataset. The GPT-4o scores out of 10 are shown for each method.
    }
    \label{fig:qualitative6}
\end{figure*}

\end{document}